\pdfoutput=1

\documentclass[11pt]{article}

\usepackage[preprint]{acl}

\usepackage{times}
\usepackage{latexsym}

\usepackage[T1]{fontenc}

\usepackage[utf8]{inputenc}

\usepackage{microtype}

\usepackage{inconsolata}

\usepackage{graphicx}

\usepackage{url}

\usepackage{CJKutf8}
\usepackage[utf8]{inputenc}
\usepackage[T1]{fontenc}

\newcommand{\adparaphrase}{\textsc{AdParaphrase}}
\newcommand{\dataset}{\textsc{AdParaphrase v2.0}}
\newcommand{\datasetvone}{\textsc{AdParaphrase v1.0}}
\newcommand{\vone}{\textsc{v1.0}}
\newcommand{\vtwo}{\textsc{v2.0}}

\usepackage{booktabs}
\usepackage{multirow}
\usepackage{amssymb}

\title{\dataset: Generating Attractive Ad Texts Using a Preference-Annotated Paraphrase Dataset}

\author{
    Soichiro Murakami$^{1}$, \ Peinan Zhang$^1$, \ Hidetaka Kamigaito$^{2,3}$, \\
    {\bf Hiroya Takamura}$^3$, \  {\bf Manabu Okumura}$^3$ \\
  $^1$CyberAgent, Inc., $^2$Nara Institute of Science and Technology, $^3$Institute of Science Tokyo \\
  {\tt 	\{murakami\_soichiro,zhang\_peinan\}@cyberagent.co.jp}, \\
  {\tt kamigaito.h@is.naist.jp}, {\tt \{takamura,oku\}@pi.titech.ac.jp} \\
}

\begin{document}
\maketitle
\begin{abstract}
Identifying factors that make ad text attractive is essential for advertising success. This study proposes \dataset, a dataset for ad text paraphrasing, containing human preference data, to enable the analysis of the linguistic factors and to support the development of methods for generating attractive ad texts. Compared with \vone, this dataset is 20 times larger, comprising 16,460 ad text paraphrase pairs, each annotated with preference data from ten evaluators, thereby enabling a more comprehensive and reliable analysis. Through the experiments, we identified multiple linguistic features of engaging ad texts that were not observed in \vone~and explored various methods for generating attractive ad texts. Furthermore, our analysis demonstrated the relationships between human preference and ad performance, and highlighted the potential of reference-free metrics based on large language models for evaluating ad text attractiveness.
The dataset is publicly available at: \url{https://github.com/CyberAgentAILab/AdParaphrase-v2.0}.\footnote{The dataset is provided under the CC BY-NC-SA 4.0 license.}
\end{abstract}

\section{Introduction\label{sec:introduction}}

Advertisements play a vital role in marketing, raising awareness of products or services, capturing user interests, and driving actions such as clicks. To maximize their effectiveness, ad writers must create attractive ad texts that appeal to users. However, with the growing demand for online advertising, manual ad text creation is reaching practical limitations, highlighting the need for automatic ad text generation (ATG) \cite{murakami2023atgsurvey}. Writing attractive ad texts requires considering two aspects: \textit{what-to-say} (the content to be advertised, such as price or product name) and \textit{how-to-say} (the way the content is expressed). This study focuses on the \textit{how-to-say} aspect in exploring methods for generating attractive ad texts, aiming to identify linguistic factors 
that capture the user's interest. 

\begin{table}[t]
\centering
{\small
\begin{tabular}{@{}clr          @{}}
\toprule
                     & \multicolumn{1}{c}{\textbf{Ad Text}}       & \textbf{\#Pref} \\ \midrule
\multirow{2}{*}{(a)} & \textit{Up to 50\% discount on your first purchase} & 0 \\
                     & \textit{Get up to 50\% off on your first purchase}  & 9 \\ \cmidrule{1-3}
\multirow{2}{*}{(b)} & \textit{The industry's lowest prices}               & 3 \\
                     & \textit{Top-class low prices in the industry}       & 7 \\ 
\bottomrule
\end{tabular}}
\caption{Example of \dataset, translated into English for visibility. \#Pref represents the number of evaluators who preferred each ad text. Those who chose ``skip'' are not included.}\label{tab:paraphrase_example}
\end{table}





Many studies have investigated the factors that influence ad performance and human preference \cite{youngmann2020,yuan2023persuadetoclick}. However, identifying the linguistic factors presents a significant challenge because of the intricate interplay between the semantic content 
and its linguistic expression. A clear analysis of the linguistic factors requires disentangling them 
and focusing exclusively on their impact \cite{pryzant-etal-2018-interpretable}.

To address this challenge, \citet{murakami2025adparaphrasev1} introduced \textsc{AdParaphrase}, which is a dataset comprising paraphrase pairs of ad texts, annotated with human preferences from ten evaluators. By controlling the content, the dataset allows us to investigate how linguistic expressions alone affect the attractiveness of the ad. Using this dataset, they 
identified linguistic factors, such as noun count, that significantly affect human preferences. In addition, they demonstrated that these findings can improve the generation of attractive ad texts.

However, the small size of their dataset, \adparaphrase, presents notable limitations. The dataset contains only 725 paraphrase pairs created by professional ad writers and is insufficient for conducting comprehensive and reliable analyses or training ATG models. Consequently, previous studies have primarily relied on in-context learning (ICL) \cite{brown_gpt3_2020}, leaving other promising approaches, such as preference tuning \cite{rafailov2024dpo}, unexplored.

To address these limitations, we present 
\dataset, an expanded version of the original dataset, 
with referring to the original dataset as \vone. 
Table~\ref{tab:paraphrase_example} presents paraphrase examples from the dataset.
The number of paraphrase pairs annotated with human preferences in \vtwo~is approximately 20 times larger than \vone. 
This expansion enables a comprehensive analysis and encourages the exploration of other ATG approaches. The dataset was built using scalable methods including large language models (LLMs) and crowdsourcing, with manual annotations for paraphrase identification 
and preference judgment.

In the experiments, we analyzed \dataset~and identified multiple linguistic factors influencing human preferences that were not identified in \vone~(\S\ref{sec:feature_analysis}). We then evaluated various methods for generating attractive ad texts, including ICL, instruction tuning, and preference tuning, by examining the characteristics of each approach (\S\ref{sec:attractive_ad_text_generation}). In addition, our analysis identified the relationships between human preferences and ad performances, and demonstrated the suitability of reference-free metrics for the automatic evaluation of ad text attractiveness (\S\ref{sec:discussion}). We hope \dataset~will drive further advancements in generating attractive ad texts.

\section{Related Work}

\subsection{Ad Text Optimization}
\looseness=-1
Optimizing ad texts to enhance ad performance is a critical challenge for advertisers. To this end, various approaches have been developed such as ATG and text analysis \cite{murakami2023atgsurvey}. 

ATG approaches are broadly classified into two categories: generation from scratch \cite{Bartz2008-ke,Hughes2019-sh} and ad text refinement \cite{youngmann2020,murakami2025adparaphrasev1}. The former involves creating ad text from sources, such as keywords and landing pages~\cite{Kamigaito2021-iy,mita-etal-acl2024-camera}, whereas the latter focuses on improving existing ad texts~\cite{mishra2020refinement}. This study falls into the latter category.

Using text analysis, previous studies investigated factors affecting attractiveness, such as persuasion strategies \cite{yuan2023persuadetoclick}, emotions \cite{youngmann2020}, and advertising appeal \cite{murakami-etal-2022-aspect}. The key difference between previous studies and our work is that we focus on the attractiveness of linguistic expression in ad texts. The factors that influence attractiveness can be broadly divided into \textit{what-to-say} and \textit{how-to-say}. Although previous studies often focused on \textit{what-to-say} without explicitly distinguishing between the two, we specifically focus on \textit{how-to-say}.

\subsection{Paraphrase Generation}
Our study is closely related to paraphrase generation, as it focuses on rephrasing ad texts into more attractive expressions while preserving their meaning. Paraphrase generation has long been a central challenge in natural language processing, with numerous datasets and methods proposed across various domains \cite{zhou-bhat-2021-paraphrase-survey}. 

This study differs from previous studies in two key aspects: First, it targets ad texts, a domain with unique characteristics distinct from previously studied areas such as social media \cite{lan-etal-2017-continuously} and questions \cite{zhang-etal-2019-paws}. Second, it prioritizes human preference in paraphrase pairs, specifically examining linguistic expressions that enhance the attractiveness of ad texts—a perspective unique to the advertising domain. We hope that our dataset will expand the scope of future research on paraphrase generation.

\section{Method of Dataset Construction\label{sec:paraphrase_dataset_construction}}
\looseness=-1
This section describes the design principles of \dataset~(\S\ref{sec:dataset_design_principles}), the three-step construction process involving paraphrase candidate collection (\S\ref{sec:collecting_paraphrase_candidates}), paraphrase identification~(\S\ref{sec:paraphrase_identification}), and preference judgment (\S\ref{sec:human_preference_judgments}), and the quality control measures implemented throughout its construction (\S\ref{sec:quality_control}).

\subsection{Principles of Dataset Design\label{sec:dataset_design_principles}}
Our design principles are threefold: (1) ensuring that the dataset is 
large enough to support both analysis and model training; (2) incorporating a diverse range of paraphrasing cases; and (3) making the dataset publicly available under a proper license for research purposes.

To achieve Principle (1), over 10,000 data samples were collected. This quantity was determined based on the benchmarks and requirements observed in previous studies \cite{jha2023limit,mita-etal-acl2024-camera} for reliable data analysis and effective model training.
To address Principle (2), a wide range of paraphrased expressions were covered beyond simple phenomena such as ``word order changes'' by providing explicit stylistic instructions during paraphrase generation.
Finally, in line with Principle (3), the dataset was constructed using methods compatible with open distribution for research purposes. Specifically, we leveraged crowdsourcing and utilized open LLMs whose licenses permit the redistribution of the generated content.

\subsection{Collecting Paraphrase Candidates \label{sec:collecting_paraphrase_candidates}}
\dataset~was constructed based on CAMERA~\cite{mita-etal-acl2024-camera}, a publicly available Japanese ad text dataset for ATG. In this study, by leveraging all ad texts from the dataset as source texts, we collected paraphrase pairs 
by generating their paraphrases using both LLMs and crowdworkers. While the quality would be ensured by relying solely on professional ad writers to create paraphrases, it is impractical to construct large-scale datasets with the method because of resource constraints. To address this issue, we leveraged 133 high-quality paraphrase pairs from \datasetvone~created by professional ad writers as references for LLMs and crowdworkers. This approach combines the expertise of professional writers with automated methods to efficiently generate numerous paraphrase candidates. The procedure for generating paraphrases using LLMs and crowdworkers is as follows:

\paragraph{Paraphrase Generation using LLMs}
Paraphrase candidates were generated using LLMs, known for their paraphrase-generation capabilities \cite{cegin-etal-2023-chatgpt}, via In-Context Learning (ICL)~\cite{brown_gpt3_2020}. 
For this approach, high-quality paraphrase examples from professional writers were provided as few-shot examples, along with instruction texts as prompts.
To enhance the diversity of paraphrases in accordance with Principle (2), stylistic instructions were also incorporated into the prompts. We defined 40 types of stylistic instructions, such as \textit{``Use simpler syntax''}, to guide LLMs in generating paraphrase candidates based on specified styles.\footnote{The results from our analysis of the effect of stylistic instructions are provided in Appendix~\ref{appendix:effect_of_stylistic_instructions}. We confirmed that explicitly specifying stylistic instructions enables the generation of lexically and syntactically diverse paraphrase candidates.}
Stylistic instructions were randomly selected for each ad text. Examples of prompts and stylistic instructions  are provided in Appendix~\ref{appendix:collecting_paraphrase_candidates}. 
Moreover, multiple LLMs with different training datasets and model sizes were used. The selection of LLMs was based on Principle (3) and whether they were pre-trained on Japanese corpora. Specifically, we selected four models.\footnote{The four models include tokyotech-llm/Llama-3.1-Swallow-8B-Instruct-v0.1, tokyotech-llm/Llama-3.1-Swallow-70B-Instruct-v0.1, cyberagent/calm2-7b-chat, and cyberagent/calm3-22b-chat on Hugging Face Hub~\cite{wolf-etal-2020-transformers}.} For example, \texttt{Swallow-70B} is a model based on Llama 3.1 and is distributed under the Llama 3.1 license,\footnote{\url{https://www.llama.com/llama3_1/license/}} which permits the use of model-generated texts for research purposes, including model training.

\paragraph{Paraphrase Generation by Crowdworkers}
We used a crowdsourcing service.\footnote{\url{https://crowdsourcing.yahoo.co.jp/}} The same instructions and paraphrase examples as those given to the LLMs were provided to the crowdworkers as annotation guidelines.
Because most workers lack experience in creating ad texts, additional knowledge about ad text creation (e.g., \textit{``Include words that encourage action''}) was also included in the guidelines. The complete guidelines are available in Appendix~\ref{appendix:collecting_paraphrase_candidates}.


\subsection{Paraphrase Identification\label{sec:paraphrase_identification}}
Manual labeling was conducted to indicate whether the generated candidates are really a paraphrase at the sentence level. 
To reduce the manual labor, we first applied rule-based filtering to exclude (1) candidates that are clearly not a paraphrase (e.g., contain different dates or monetary amounts) and (2) ad texts exceeding 30 characters. The length constraint was based on guidelines from ad platforms such as Google Ads\footnote{\url{https://ads.google.com}} because texts beyond this limit cannot be delivered. Paraphrase identification (PI) was conducted via crowdsourcing, whereby five workers evaluated each ad text pair and made a binary judgment on whether it qualifies as a paraphrase. The final label for each pair was determined by majority vote. 
The instructions provided to the workers and example paraphrase pairs are presented in Appendix~\ref{appendix:paraphrase_identification} and \ref{appendix:example_paraphrase_pairs}, respectively.
 
\subsection{Human Preference Judgment\label{sec:human_preference_judgments}}
Preference judgments were conducted for valid paraphrase pairs via crowdsourcing, with each pair judged by ten workers. Workers were asked to select the more attractive ad text or ``skip'' if both were equally attractive. To address the subjective nature of preference judgments, we followed the guidelines of~\citet{Wang2021-uq} and provided the workers with multiple aspects of attractiveness, such as \textit{``more clickable?''} and \textit{``easier to understand''} as well. The complete annotation guidelines are provided in Appendix~\ref{appendix:attractiveness_eval}.

\subsection{Quality Control\label{sec:quality_control}}
Several measures were implemented to ensure high annotation quality despite inherent biases, such as positional bias \cite{wang-etal-2024-large-language-models-fair}.
Positional bias was mitigated by randomizing the order of the options presented to the workers. 
In addition, attention checks \cite{klie-etal-2024-analyzing} were included using identical ad text pairs with predefined correct answers (e.g., \textit{paraphrase} for the PI task and \textit{skip} for preference judgment), rejecting responses from annotators failing these checks to maintain quality.

\section{Dataset Statistics and Analysis\label{sec:analysis_of_adparaphrase_v2}}
\subsection{Dataset Statistics}

\begin{table}[t]
\centering
\setlength{\tabcolsep}{3.5pt}
{\small
\begin{tabular}{@{}l@{\hspace{-1em}}rrr|rr@{}}
\toprule
\multicolumn{1}{c}{\multirow{2}{*}{\textbf{Model}}} & 
\multicolumn{1}{c}{\multirow{2}{*}{\textbf{\#Generated}}}& 
\multicolumn{1}{c}{\multirow{2}{*}{\textbf{\#Filtered}}}& 
\multicolumn{1}{c}{\multirow{2}{*}{\textbf{\#Para.}}} &
\multicolumn{2}{c}{\textbf{Pass Rates}} \\
& & & & \multicolumn{1}{c}{\textbf{PI}} & \multicolumn{1}{c}{\textbf{Pref}} \\
\midrule
\texttt{CALM2-7B}    & 16,365 & 2,107  & 1,173  & 7.2   & 22.9 \\
\texttt{CALM3-22B}   & 16,365 & 6,287  & 4,551  & 27.8  & 21.4 \\
\texttt{Swallow-8B}  & 16,365 & 4,942  & 3,623  & 22.1  & 20.9 \\
\texttt{Swallow-70B} & 16,365 & 5,226  & 4,174  & 25.5  & 19.5 \\
Crowd worker & 5,000  & 3,775  & 2,939  & \textbf{58.8}  & \textbf{25.8} \\ \midrule
\multicolumn{1}{r}{\textbf{Total}}          & 70,460 & 22,337 & 16,460 & 23.4  & 21.7 \\

\bottomrule
\end{tabular}}
\caption{Statistics of \dataset. \#Generated, \#Filtered, \#Para. refer to the number of generated paraphrase candidates, the number of paraphrase candidates that passed the rule-based filtering, and the number of valid paraphrases judged by a majority of workers, respectively. PI and Pref stand for the pass rates of paraphrase identification and preference judgment. \label{tab:paraphrase_dataset_statistics}}
\end{table}

Table \ref{tab:paraphrase_dataset_statistics} summarizes the dataset statistics obtained for the paraphrase construction process described in \S\ref{sec:paraphrase_dataset_construction}.
First, for paraphrase candidate collection, 16,365 ad texts from CAMERA were used as inputs, obtaining 70,460 paraphrase candidates through four LLMs and crowdsourcing. As source text, crowdworkers used 5,000 texts randomly sampled from CAMERA.
Second, rule-based filtering was applied, resulting in 22,337 paraphrase candidates. Many candidates were removed during this filtering step primarily because they exceeded the length constraints.
Third, 16,460 candidates were judged as paraphrases in PI.
Finally, conducting preference judgments on the identified paraphrase pairs yielded 16,460 pairs of preference judgment data.

\subsection{Inter-Annotator Agreement\label{sec:inter_annotator_agreement}}
Inter-annotator agreement (IAA) for PI (\S\ref{sec:paraphrase_identification}) and preference judgment (\S\ref{sec:human_preference_judgments}) was measured using Fleiss' kappa~\cite{fleiss1971mns}. The kappa value for PI was 0.442, indicating moderate agreement, whereas that for preference judgment was 0.167, indicating slight agreement~\cite{richard1977iaa}. The relatively low agreement in preference judgment likely reflects the subjective nature, which is consistent with the results of previous studies on ad text evaluation \cite{mita-etal-acl2024-camera}.

\subsection{Evaluation of Paraphrase Candidates\label{sec:evaluation_of_generated_paraphrase_candidates}}
Table~\ref{tab:paraphrase_dataset_statistics} presents the pass rates for PI and preference judgment across different models. The pass rate for PI represents the proportion of generated texts that passed both rule-based filtering and manual annotation, whereas the pass rate for preference judgment indicates the proportion of paraphrases judged as attractive by at least eight evaluators. 

For PI, crowdworkers achieved the highest pass rate, and larger LLMs such as CALM3-22B performed better. 
In preference judgment, crowdworkers again outperformed LLMs, with 25.8\% of their paraphrases judged as attractive. Among LLMs, CALM2-7B showed a slightly higher rate. The gap between LLMs and crowdworkers in preference judgment was small, suggesting that LLMs, despite slightly underperforming humans, are still effective for generating attractive paraphrases.

\subsection{Distribution of Preference Judgments\label{sec:distribution_of_preference_judgments}}
\begin{figure}[t]
 \centering
  \includegraphics[width=1\linewidth]{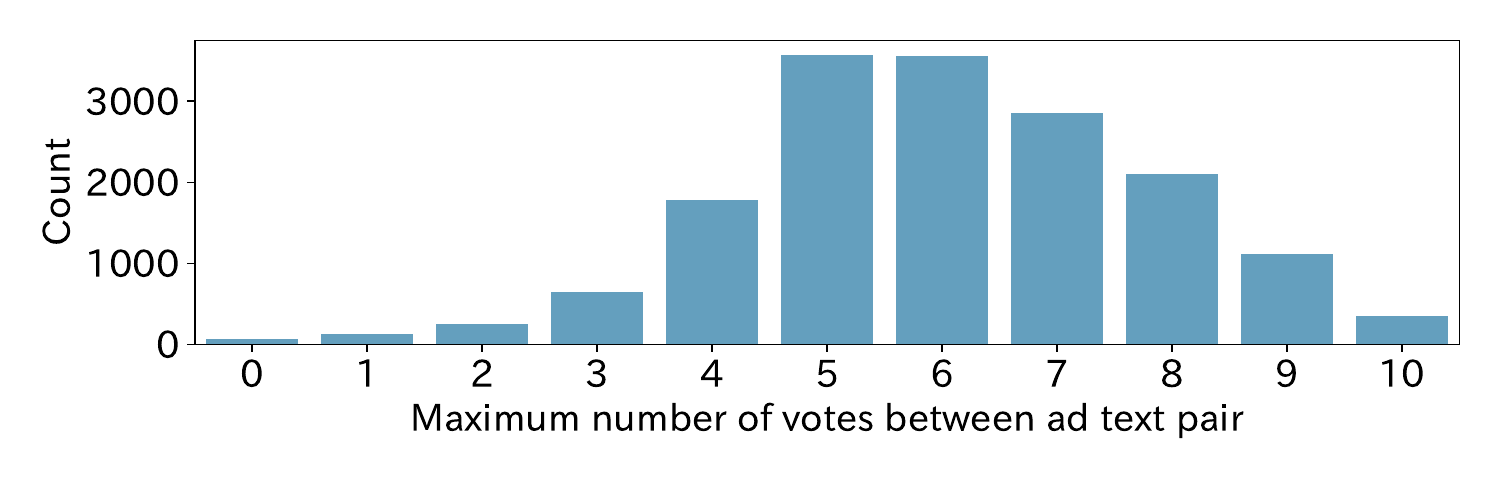}
 \caption{Distribution of maximum number of votes between ad text pair in preference judgments.}
 \label{fig:dist_of_attractiveness_evaluation}
\end{figure}
\looseness=-1
The histogram showing the distribution of preference judgment results is presented in Figure~\ref{fig:dist_of_attractiveness_evaluation}. The x-axis represents the number of evaluators who preferred the same ad text, excluding ``skip'' responses. For example, a value of seven indicates that seven out of ten evaluators preferred the same ad text, whereas zero indicates that all evaluators skipped it. 

The distribution of preference judgments and their IAA (\S\ref{sec:inter_annotator_agreement}) revealed an inconsistency in human preference for ad text paraphrase pairs. Specifically, the most common agreement level involved five to six evaluators. 
However, 3,570 cases, with at least eight evaluators preferring the same ad text, showed moderate agreement with an IAA of 0.480, measured by Fleiss' kappa.
This non-random agreement level, which was particularly noticeable in cases of strong preference, suggests that differences in linguistic expressions are likely to influence human preferences.

\subsection{Dataset Comparison}
\looseness=-1
\begin{table}[t]
\setlength{\tabcolsep}{3.5pt}
\centering
{\small
\begin{tabular}{@{ }clrccc@{ }}
\toprule
\textbf{Ver.} & \multicolumn{1}{c}{\textbf{Labels}} & \multicolumn{1}{c}{\textbf{\#Pairs}} & \textbf{Pref.} & \textbf{Training} & \textbf{Creators} \\ \midrule
\multirow{2}{*}{\vtwo} & Para     & 16,460 & $\checkmark$ & \multirow{2}{*}{Allowed} & Crowdworker, \\
                       & Non-Para & 5,877  & $-$          &                          & Open LLMs     \\ \midrule
\multirow{2}{*}{\vone} & Para     & 725    & $\checkmark$ & \multirow{2}{*}{Limited} & Ad writers,   \\
                       & Non-Para & 513    & $-$          &                          & Closed LLMs   \\ \midrule
\end{tabular}
\caption{Comparison of \datasetvone~and \vtwo.\label{tab:comparison_with_v1}}}
\end{table}
Table~\ref{tab:comparison_with_v1} compares \datasetvone~and \vtwo. 
\dataset~includes over 20 times more paraphrases compared to \vone. Furthermore, our dataset adheres to Principle (3), in that it is freely available for research, including model training. In contrast, \vone~relies on GPT-3.5 and GPT-4 via the Azure OpenAI API, that imposes licensing restrictions that limit its usability.\footnote{\url{https://azure.microsoft.com/en-us/products/ai-services/openai-service/}}

\section{Experiments}
Through dataset construction, we collected ad text pairs with human preference annotations that were 20 times larger in scale than those in \vone. 
Using the dataset, we conducted two experiments: (1) an analysis of linguistic features influencing human preferences and (2) an ATG task. The first experiment leveraged our larger dataset to identify the linguistic features influencing human preferences that were not revealed in \vone. The second experiment evaluated the effectiveness of recent text-generation techniques, such as instruction tuning \cite{wei2022instructiontuning} and preference tuning \cite{rafailov2024dpo}, for the ATG task. This extends the prior work limited to ICL. Through the experiment, we assessed the potential of these methods for generating more attractive ad texts.

\subsection{Analysis of Linguistic Features \label{sec:feature_analysis}}
In this experiment, we focused on 3,570 paraphrase pairs with moderate preference agreement (\S\ref{sec:distribution_of_preference_judgments}), analyzing how differences in linguistic expressions influence preferences using a chi-square test.

\subsubsection{Experimental Settings\label{sec:experimental_settings_for_linguistic_feature_analysis}}
\paragraph{Linguistic Features}
The objective of ad texts is to capture people's attention and draw their interest. Thus, factors such as visibility, informativeness, and readability play a crucial role in enhancing their attractiveness~\cite{hsuehcheng_wang__2012,schwab2013how-to-write-a-good-advertisement}. We analyzed how linguistic features related to expression and style influence human preferences. Following \citet{murakami2025adparaphrasev1}, linguistic features were categorized into four groups: \textit{raw text}, \textit{lexical}, \textit{syntactic}, and \textit{stylistic}. A list of the linguistic features is presented in Table~\ref{tab:chi_square_test_limited}.\footnote{Only a subset of features is presented in Table~\ref{tab:chi_square_test_limited} due to space limitations. The complete list of 26 features, along with their definitions and analysis results, is in Appendix~\ref{appendix:linguistic_features}.}
As a raw text feature, we used character count, which affects the informativeness and readability of the text.
The lexical features include the number of content words, character types, and lexical choice. Content words are related to informativeness, whereas character types are associated with readability \cite{sato-etal-2008-automatic}. Lexical choice was measured by counting common and proper nouns, assuming that commonly used words are preferred.
Syntactic features measure text complexity and fluency, including the depth of the dependency tree, the dependency link length, and perplexity (PPL).
Stylistic features include emotion, textual specificity, and decorative use of symbols. The emotion and specificity labels were assigned using external classifiers, as described in Appendix \ref{appendix:linguistic_features}. For decorative symbols, the presence of brackets was included, as they are widely used in Japanese ad texts to emphasize key information. 

\paragraph{Analysis Method}
To analyze the relationship between each linguistic feature and human preference, we used the chi-square test of independence. This method assesses the independence between two categorical variables: (1) ad texts preferred by most evaluators and (2) superiority or inferiority of each linguistic feature. For example, when studying PPL, the relationship between preferred ad texts and their perplexity scores is analyzed. 

\paragraph{Dataset}
We used 3,570 ad text pairs for which at least eight out of ten evaluators expressed a preference (\S\ref{sec:human_preference_judgments}), ensuring the reliability of the factor analysis influencing preferences. In addition, to focus on the differences in linguistic expressions between ad text pairs, we analyzed only the pairs with different scores for linguistic features, such that the number of cases for each feature varied. For example, 2,925 pairs had different character counts.

\subsubsection{Results\label{sec:feature_analysis_results}}

\begin{table}[t]
\centering
\setlength{\tabcolsep}{1.9pt}
{\small
\begin{tabular}{@{}c@{\hspace{0.5em}}lrrr@{\hspace{1.1em}}r@{}}
\toprule
\multicolumn{2}{c}{\textbf{Linguistic Features}} & 
\multicolumn{1}{c}{\textbf{df}} & 
\multicolumn{1}{c}{\textbf{N}} & 
\multicolumn{1}{c}{\textbf{$\chi^2$}} & 
\multicolumn{1}{c}{\textbf{$\phi$}} \\ \midrule 
\multirow{2}{*}{\shortstack{Raw text\\features}}  
    & \textit{Text length} & & & & \\
    & \hspace{1em} character$^{\dag,\ddag,\uparrow,\ast}$      & 1  & 2,925 & 723.8  & 0.497 \\ 
\midrule
\multirow{10}{*}{\shortstack{Lexical\\features}}  
    & \textit{Content words} & & & & \\
    & \hspace{1em} noun$^{\dag,\ddag,\uparrow,\ast}$           & 1  & 1,406 & 326.6  & 0.482 \\ 
    & \hspace{1em} verb$^{\ddag,\downarrow,\ast}$              & 1  & 535   & 6.9    & 0.114 \\ 
    & \hspace{1em} adjective                                   & 1  & 99    & 0.9    & 0.094 \\ 
    & \textit{Lexical choice} & & & & \\
    & \hspace{1em} common noun$^{\dag,\ddag,\uparrow,\ast}$    & 1  & 1,397 & 288.1  & 0.454 \\ 
    & \hspace{1em} proper noun$^{\ddag,\uparrow,\ast}$         & 1  & 152   & 7.6    & 0.223 \\ 
    & \textit{Character type} & & & & \\
    & \hspace{1em} hiragana$^{\ddag,\downarrow,\ast}$          & 1  & 2,047 & 23.2   & 0.107 \\ 
    & \hspace{1em} kanji$^{\ddag,\uparrow,\ast}$               & 1  & 1,503 & 257.7  & 0.414 \\ 
\midrule                                
\multirow{6}{*}{\shortstack{Syntactic\\features}} 
    & \textit{Dependency tree} & & & & \\
    & \hspace{1em} depth$^{\ddag,\downarrow,\ast}$             & 1  & 1,914 & 16.9   & 0.094 \\ 
    & \hspace{1em} length                                      & 1  & 2,349 & 1.9    & 0.028 \\ 
    & \textit{Others} & & & & \\
    & \hspace{1em} noun phrases$^{\dag,\ddag,\uparrow,\ast}$   & 1  & 1,895 & 259.8  & 0.370 \\ 
    & \hspace{1em} perplexity$^{\dag,\ddag,\downarrow,\ast}$   & 1  & 3,570 & 223.3  & 0.250 \\ 
\midrule
\multirow{6}{*}{\shortstack{Stylistic\\features}} 
    & \textit{Emotion} & & & & \\
    & \hspace{1em} joy$^{\ddag,\downarrow,\ast}$               & 1  & 693   & 70.1   & 0.318 \\ 
    & \hspace{1em} anticipation$^{\ddag,\uparrow,\ast}$        & 1  & 683   & 89.3   & 0.362 \\ 
    & \textit{Others} & & & & \\
    & \hspace{1em} specificity$^{\ddag,\uparrow,\ast}$         & 1  & 186   & 116.4  & 0.791 \\ 
    & \hspace{1em} brackets$^{\dag,\ddag,\uparrow,\ast}$       & 1  & 1,667 & 1,372.6& 0.907 \\ 
\bottomrule                                
\end{tabular}}

\caption{Results of the chi-square test. Df, N, and $\phi$ refer to the degree of freedom, the number of cases for each feature, and the measure of effect size, respectively. \ddag~indicates linguistic features, identified in \vtwo, that influence preference judgments, while \dag~denotes those identified in \vone. $\uparrow$ and $\downarrow$ indicate that ad texts with higher and lower feature scores, respectively, are preferred. $\ast$ indicates a significant relationship with human preferences ($p<0.01$).}\label{tab:chi_square_test_limited}
\end{table}
Table~\ref{tab:chi_square_test_limited} presents the chi-square test results. Linguistic features with higher chi-square values ($\chi^2$) and lower p-values indicate a stronger relationship with human preferences.
We also report Phi ($\phi$), a commonly used measure of effect size for the chi-square test~\cite{cohen1988statisticalpower}.
$\phi$ is defined as $\sqrt{(\chi^2/N)}$, where $N$ is the number of observations. A value of 0.1 is considered a small effect, 0.3 a medium effect, and 0.5 a large effect.

These results reveal that several linguistic features, such as textual specificity and certain emotions (e.g., joy, anticipation), which were not identified by \vone, are significantly related to human preferences. 
Specifically, cross-tabulations between linguistic features and preference judgments showed that ad texts with the following characteristics were preferred: \textit{longer text}, \textit{more nouns}, \textit{shallower dependency trees}, \textit{lower perplexity}, \textit{higher specificity}, and the \textit{presence of brackets}. These are examples of preferred features, and the full results are presented in Appendix~\ref{appendix:linguistic_features}. Conversely, no significant differences were observed for features such as the number of adjectives.

\subsection{Ad Text Generation\label{sec:attractive_ad_text_generation}}
In this experiment, we focus on ad text refinement \cite{mishra2020refinement}, a task that generates more attractive ad texts by rephrasing the linguistic expressions without adding or removing any information. 

\subsubsection{Experimental Settings\label{sec:experimental_settings_for_atg_generation_methods}}
\paragraph{Comparison Methods}
In exploring multiple methods for generating more attractive ad texts, we focused on recent LLM-based techniques, such as instruction tuning \cite{wei2022instructiontuning}, preference tuning \cite{rafailov2024dpo}, and ICL \cite{brown_gpt3_2020}. For ICL, we tested three types of prompts: (1) \texttt{zeroshot}, which provides only basic instructions for rephrasing an input ad text into a more attractive ad text; (2) \texttt{zeroshot-findings}, which further incorporates feature analysis findings (in \S\ref{sec:feature_analysis}) into the prompt; and (3) \texttt{fewshot-findings}, which extends \texttt{zeroshot-findings} by including 20 paraphrase examples sampled from the training data. 
As the findings, we incorporated higher character counts, greater fluency, and the use of brackets into the prompt. The few-shot examples were selected based on preference judgments, pairing less-preferred input texts with their corresponding preferred outputs.
For instruction tuning, the LLMs were fine-tuned using less-preferred ad texts as inputs and highly preferred ad texts as outputs, based on human preference judgments. The instruction-tuned models were further refined by preference tuning via direct preference optimization (DPO) \cite{rafailov2024dpo}. For further implementation details, including the training setups and prompts used for each model, please refer to Appendix \ref{appendix:ad_text_generation}.

\paragraph{LLMs}
Three LLMs, CALM3-22B~\cite{cyberagent-calm3-22b-chat}, Swallow70B~\cite{fujii2024swallow}, and GPT-4o~\cite{openai2024gpt4o}, were employed. The first two models were chosen because they were pre-trained on Japanese corpora, either from scratch or through continual learning. We used GPT-4o via the Azure OpenAI API, version \texttt{2024-09-01-preview}. Additionally, to compare the human performance with those of LLMs, the paraphrases created by crowdworkers were evaluated. Crowdworkers were instructed to create paraphrases from the given ad text based on the guidelines described in \S\ref{sec:collecting_paraphrase_candidates}.
 
\paragraph{Dataset}
A revised version of \dataset~was used for model training.\footnote{In AdParaphrase v2.0, preference judgments were conducted on ($x, y$). However, this data format is not suitable for preference tuning such as DPO. Thus, the triplets ($x, y_1, y_2$) were created, and preference data were collected for ($y_1, y_2$).} Specifically, the triplets ($x$, $y_1$, $y_2$) were formed by pairing source ad text $x$ and two paraphrases $y_1$ and $y_2$ generated by the different models in \vtwo. 
Subsequently, preference judgments were conducted for $y_1$ and $y_2$ using the annotation process in \S\ref{sec:human_preference_judgments}, collecting responses from ten evaluators. As a result, we constructed a dataset of 8,721 triplets ($x$, $y_1^\mathrm{pref}$, $y_2^\mathrm{pref}$), where $y_1^\mathrm{pref}$ and $y_2^\mathrm{pref}$ denote preference-labeled paraphrases. 
The dataset was split into training, development, and test sets at a ratio of $9:0.5:0.5$.

\paragraph{Evaluation Methods}
\begin{table}[t]
\setlength{\tabcolsep}{4pt}
\centering
{\small
\begin{tabular}{@{ }lrrr@{ }}
\toprule
\multicolumn{1}{c}{\textbf{Model}} & 
\multicolumn{1}{c}{\textbf{PI}} & 
\multicolumn{1}{c}{\textbf{Att}} & 
\multicolumn{1}{c}{\textbf{Att\&Length}} \\ \midrule

CALM3-22B & & & \\
\hspace{1em} zeroshot          & 74.0          & 23.0          & 12.8            \\
\hspace{1em} zeroshot-findings & 74.0          & 42.6          & 23.0            \\
\hspace{1em} fewshot-findings  & 85.0          & 38.8          & \textbf{31.2}   \\
\hspace{1em} instruct-zeroshot & \textbf{90.5} & 31.5          & 29.3            \\
\hspace{1em} dpo-zeroshot      & 70.5          & \textbf{84.4} &  8.5            \\ \midrule

Swallow70B & & & \\
\hspace{1em}zeroshot           & 90.5          & 15.5          &  8.3            \\
\hspace{1em}zeroshot-findings  & 80.0          & 44.4          & 17.5            \\
\hspace{1em}fewshot-findings   & 86.5          & 40.5          & \textbf{26.0}   \\
\hspace{1em}instruct-zeroshot  & \textbf{94.0} & 18.6          & 17.6            \\
\hspace{1em}dpo-zeroshot       & 62.5          & \textbf{71.2} &  8.0            \\ \midrule

GPT-4o & & & \\
\hspace{1em}zeroshot           & 86.0          & 12.8          & 12.8            \\
\hspace{1em}zeroshot-findings  & \textbf{95.5} & \textbf{39.3} & \textbf{34.6}   \\
\hspace{1em}fewshot-findings   & 92.5          & 37.8          & 33.5            \\ \midrule
Crowdworker                    & 89.1           & 23.9         & 22.3            \\ \bottomrule





\end{tabular}}
\caption{Human evaluation results of ATG experiments. The evaluation used three metrics: PI, Att, and Att\&Length, denoting the pass rate for paraphrase identification, the pass rate for attractiveness judgment, and the pass rate for attractiveness when length constraints are also considered, respectively.\label{tab:ad_text_generation_results}}
\end{table}
The generated texts were evaluated using three criteria: (1) paraphrase identification, (2) attractiveness, and (3) attractive while satisfying length constraints. Criteria (1) and (2) were assessed using the human evaluations described in \S\ref{sec:paraphrase_identification} and \S\ref{sec:human_preference_judgments}. For (1), the percentage of generated texts judged as paraphrases by the majority of evaluators was calculated. For (2), among the texts judged as paraphrases, we reported the percentage judged as attractive by the majority. This evaluates the ability to generate an ad text that is both a valid paraphrase and attractive. For (3), among the texts judged as paraphrases, the percentage judged as attractive and satisfying the length constraint of 30 characters was determined. As ad texts that exceed length constraints cannot be delivered in online advertising, this metric evaluates the practical capability of generating attractive ad texts within the length constraint.

\subsubsection{Results\label{sec:atg_evaluation_results}}
The evaluation results are presented in Table~\ref{tab:ad_text_generation_results}.
For paraphrasing, the instruction-tuned methods demonstrated better performance.
In terms of attractiveness, DPO-based models performed best overall.
Furthermore, \texttt{zeroshot-findings} and \texttt{fewshot-findings}, which incorporate the findings of linguistic feature analysis, generated more attractive texts than \texttt{zeroshot}.
This demonstrates that the findings obtained from the analysis contributed to improving the attractiveness of the generated texts. When considering attractiveness in conjunction with length constraints, the \texttt{zeroshot-findings} outperformed DPO-based models.
This is because DPO-based models generated many texts that failed the length constraint, thereby reducing their score in this comparison.

\begin{table}[t]
\setlength{\tabcolsep}{3pt}
\centering

{\small
\begin{tabular}{@{}l@{\hspace{-1em}}rrrr}
\toprule
\multicolumn{1}{c}{\textbf{Model}} & 
\multicolumn{1}{c}{\textbf{Perplexity$\downarrow$}} & 
\multicolumn{1}{c}{\textbf{\#Char$\uparrow$}} & 
\multicolumn{1}{c}{\textbf{Brackets$\uparrow$}} \\ \midrule
CALM3-22B & & & & \\
\hspace{1em} zeroshot          & 155.6          & 27.5           & 5.0            \\
\hspace{1em} zeroshot-findings & 157.6          & 30.7           & 64.5           \\
\hspace{1em} fewshot-findings  & 146.7          & 27.0           & \textbf{69.0}  \\
\hspace{1em} instruct-zeroshot & 168.5          & 24.1           & 48.5           \\
\hspace{1em} dpo-zeroshot      & \textbf{92.2}  & \textbf{42.3}  & 37.0           \\ \midrule

Crowdworker                    & 264.3 & 23.8 & 45.8 \\ \midrule
\multicolumn{1}{r}{\textbf{Source ad texts}} & 169.7 & 23.6 & 39.5 \\ \bottomrule
\end{tabular}}
\caption{Linguistic features of generated ad texts. \#Char and Brackets denote the average number of characters per text and the proportion of generated texts that include the bracket symbol, respectively.\label{tab:analysis_of_generated_texts_in_atg_experiment_limited}}
\end{table}

 
 
Table~\ref{tab:analysis_of_generated_texts_in_atg_experiment_limited} presents the linguistic features of the generated texts, including PPL, character count, and the presence of brackets, which were the key features incorporated into the prompt. The results indicate that models with higher attractiveness scores in Table \ref{tab:ad_text_generation_results} performed better across these linguistic features. 
Notably, DPO-based models exhibited higher character count.
This suggests that DPO-based models tend to generate longer texts, potentially benefiting from length heuristics~\cite{park-etal-2024-disentangling}, a bias where evaluators tend to perceive longer texts as more attractive.

\section{Analysis\label{sec:discussion}}
In this section, we report on the analyses conducted from two main perspectives, to contribute to the future development of attractive ad text generation. The first is an analysis of the relationship between human preferences and ad performance. Given that the ultimate goal of advertising is to optimize ad performance (e.g., clicks), clarifying the relationship between ad text preferences and ad performance is crucial. The second perspective concerns automatic evaluation of PI and attractiveness. Although PI and attractiveness were evaluated manually in this study, verifying automatic evaluation metrics as alternatives to manual evaluation is required to enhance efficiency in future research.

For the former perspective, we conducted two experiments: evaluating the relationship between human preference and predicted CTR (pCTR) (\S\ref{sec:relationship_between_pctr_and_human_preferences}) and assessing ad performance in a real-world environment online (\S\ref{sec:online_evaluation}). For the latter, a meta-evaluation was performed to assess the relationship between human evaluation and existing automatic evaluation metrics (\S\ref{sec:meta_evaluation}).

\subsection{Relationship between Human Preferences and pCTR\label{sec:relationship_between_pctr_and_human_preferences}}
\begin{figure}[t]
 \centering
  \includegraphics[width=1.0\linewidth]{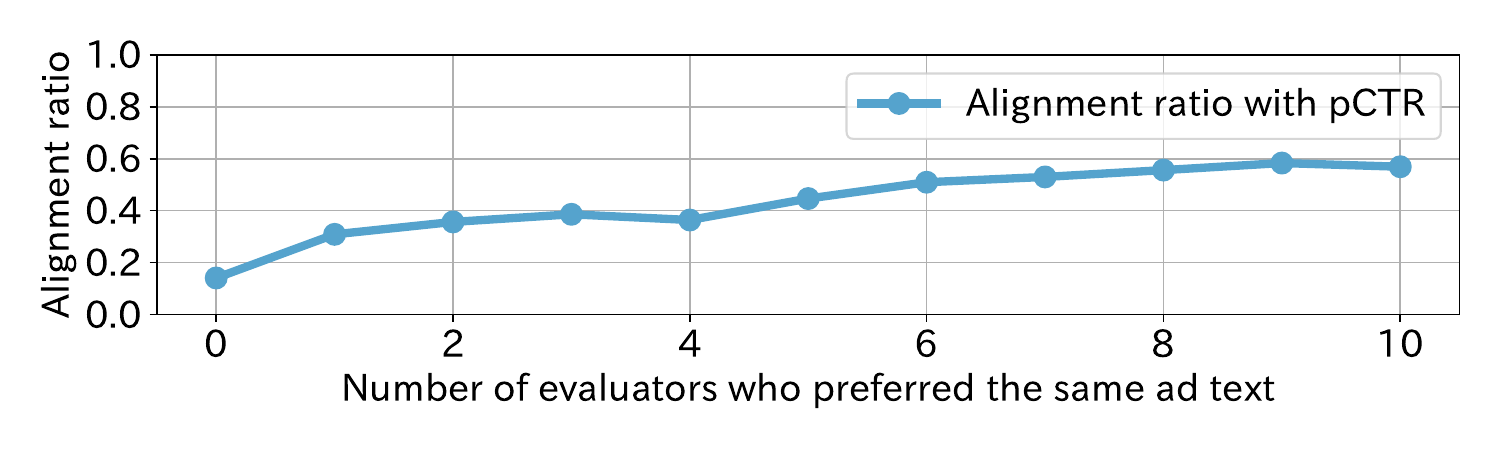}
 \caption{Alignment between human preferences and predicted Click-Through Rate (pCTR). The x-axis indicates human agreement level (number of evaluators with same preference). The y-axis shows the alignment ratio with pCTR. Higher human agreement correlates with increased alignment ratio, suggesting stronger consensus means better alignment.\label{fig:alignment_ratio_between_pctr_and_preferences}}
\end{figure}
It is critical to understand how the attractiveness of ad texts influences user behavior because the goal of advertising is to capture attentions and drive actions such as clicks. To explore this, we analyzed the relationship between human preferences and ad performance. 
Specifically, we examined whether the ad texts preferred by most evaluators also achieved a higher pCTR, a proxy for CTR.

Figure~\ref{fig:alignment_ratio_between_pctr_and_preferences} shows the alignment between human preference and pCTR in \dataset. The pCTR for each ad text was obtained using an in-house CTR prediction model. The x-axis represents the number of evaluators who preferred the same ad text, whereas the y-axis denotes the percentage of cases with pCTR and human preferences in agreement. For example, an x-axis value of ten means all evaluators preferred the same ad text in a pair, and the corresponding y-axis value shows the percentage of cases which also have higher pCTR. The results revealed a strong correlation between human preferences and pCTR (Pearson's correlation coefficient: 0.946), confirming that the ad texts preferred by the majority achieved higher CTRs. However, even when all the evaluators agreed on their preferences, the percentage of cases with a higher pCTR was approximately 60\%, suggesting a potential upper limit for improving ad performance.

\subsection{Online Evaluation of Ad Performance \label{sec:online_evaluation}}
\begin{table}[t]
\centering
\setlength{\tabcolsep}{1.8pt}
{\small
\begin{tabular}{@{}lrrrrrr@{}}
\toprule
\multicolumn{2}{c}{\textbf{Ad delivery period}}
& \multicolumn{1}{l}{\textbf{CTR}$\uparrow$}
& \multicolumn{1}{l}{\textbf{CVR}$\uparrow$}
& \multicolumn{1}{l}{\textbf{CTVR}$\uparrow$} 
& \multicolumn{1}{l}{\textbf{CPC}$\downarrow$} 
& \multicolumn{1}{l}{\textbf{CPA}$\downarrow$} \\ \midrule
Fitness                    &  2 weeks  & \textbf{91.5} & \textbf{141.4} & \textbf{129.4} & 110.7 & 79.4 \\ \midrule 
\multirow{2}{*}{Education} &  2 weeks  & 77.7 & 249.0 & 200.0 & 100.5 & 40.4 \\ 
                           &  1 month  & 93.4 & 138.9 & 127.3 & 90.6  & 65.3 \\ 
\bottomrule 
\end{tabular}}

\caption{Relative improvement of advertising performance metrics for different ad types (Fitness, Education) and delivery periods, compared to a baseline (100\%). Bold values indicate statistically significant differences, as determined by a z-test ($p<0.01$).}\label{table:online_evaluation}
\end{table}

In the online evaluation, we analyzed whether rephrasing ad texts into more attractive expressions influences ad performance, such as CTR. Specifically, we conducted an A/B test, comparing an existing group of ad texts with paraphrased ads generated using the \texttt{fewshot-findings} method\footnote{For this evaluation, we used GPT-4 as the model.} in \S\ref{sec:attractive_ad_text_generation}. The tests were conducted on Google Ads, focusing on the headline text for ads from two companies in the fitness and education industries. The ads for the former ran for two weeks, whereas those for the latter ran for two weeks or one month. Details of the evaluation setup are provided in Appendix \ref{appendix:online_evaluation}. 

Table~\ref{table:online_evaluation} summarizes the relative improvement rates of paraphrased ads over existing ads using metrics such as CTR, conversion rate (CVR), CTVR, cost per click (CPC), and cost per action (CPA).\footnote{For advertising terms, see \url{https://support.google.com/google-ads/topic/3121777}.} Among these, CTVR, defined as the product of CTR and CVR, is a comprehensive indicator of ad performance. The results indicate that as CTR decreases, CVR improves, reflecting actions such as purchases or sign-ups. Notably, for fitness ads, relative improvements in CVR and CTVR were statistically significant compared to the baseline.

\subsection{Reliability of Automatic Evaluation Metrics\label{sec:meta_evaluation}}
\begin{table}[t]
\setlength{\tabcolsep}{3.5pt}
{\small
\centering
\begin{tabular}{@{ }l@{ }rrrrrr@{ }}
\toprule
 & \multicolumn{3}{c}{\textbf{Paraphrase}} 
 & \multicolumn{3}{c}{\textbf{Attractiveness}} \\ \cmidrule(lr){2-4} \cmidrule(lr){5-7}
\textbf{Metrics}
& \multicolumn{1}{c}{$r$} 
& \multicolumn{1}{c}{$\rho$} 
& \multicolumn{1}{c}{$\tau$} 
& \multicolumn{1}{c}{$r$} 
& \multicolumn{1}{c}{$\rho$} 
& \multicolumn{1}{c}{$\tau$} \\ \midrule
BLEU     & \textbf{0.948} & 0.950          & 0.831         & -0.707         &-0.484          & -0.410         \\
ROUGE-1    & 0.279          & 0.277          & 0.199         & 0.138          & 0.204          & 0.065          \\
ROUGE-2    & 0.162          & 0.275          & 0.167         & 0.061          & -0.113         & -0.110         \\
ROUGE-L    & 0.239          & 0.306          & 0.260         & 0.197          & 0.159          & 0.051          \\
BERTScore  & 0.927          & 0.934          & 0.805         & -0.769         &-0.511          & -0.385         \\
GPT-4o     & \textbf{0.948} & \textbf{0.965} & \textbf{0.895}& \textbf{0.886} & \textbf{0.758}  & \textbf{0.615} \\ \bottomrule
\end{tabular}}
\caption{System-level meta-evaluation results with Pearson ($r$), Spearman ($\rho$), and Kendall ($\tau$).\label{tab:meta_evaluation}}
\end{table}
Adopting automatic evaluation methods is essential for enhancing efficiency in future studies. Thus, we analyzed whether existing automatic evaluation metrics can substitute human evaluations by conducting a system-level meta-evaluation. Specifically, we examined the correlations between human evaluation results from the ATG experiments (\S\ref{sec:attractive_ad_text_generation}) and various automatic metrics. The evaluation metrics are presented in Table~\ref{tab:meta_evaluation}. Inspired by the LLM-as-a-judge paradigm~\cite{gu2025surveyllmasajudge}, we included LLM-based evaluations using GPT-4o. For GPT-4o, we used human evaluation guidelines for PI and preference judgment as prompts. The LLM-based evaluation was reference-free, whereas the other metrics were reference-based, using human-created paraphrases (\S\ref{sec:attractive_ad_text_generation}) as reference texts. The automatic evaluation scores are provided in Appendix~\ref{appendix:ad_text_generation}. 

Table~\ref{tab:meta_evaluation} presents the correlations between the scores and human evaluation results. For PI, BLEU, BERTScore, and GPT-4o exhibited strong positive correlations with human evaluations. With regard to attractiveness, GPT-4o showed a strong positive correlation, whereas BLEU and BERTScore displayed negative correlations. These results suggest that both reference-based and reference-free metrics are effective in predicting PI. However, reference-free metrics are more suitable for assessing attractiveness.

\section{Conclusion}
This study introduced \dataset, a dataset for ad text paraphrasing that contains human preference data. Compared to \vone, our dataset is 20 times larger, enabling a comprehensive analysis of the key features that make ad text attractive. We identified multiple linguistic features that contribute to engaging ad texts and investigated various methods for generating attractive ad texts. Our analysis revealed the relationship between human preference and ad performance, and demonstrated the potential of reference-free evaluation for assessing ad text attractiveness. 

Future work will include enhancing ATG methods by addressing challenges such as adhering to length constraints, optimizing both human preference and ad performance, and investigating the influence of other factors on preferences, such as demographic information and product category.

\section{Limitations\label{sec:limitations}}
This study has several limitations that should be addressed in future studies.

\paragraph{Many Paraphrased Texts are LLM-Generated}
Many paraphrased texts are generated by LLMs, potentially resulting in linguistic features that differ from real ad texts. 
However, please note that the original CAMERA ads, used as source ad texts, were actually distributed ads, and so not all texts are LLM-generated.
Future research could examine expression differences between human-written and LLM-generated ads or analyze how linguistic features influence preferences, focusing on human-authored texts.

\paragraph{Language-Specific Features and Generalizability}
\dataset~is based on Japanese ad texts, meaning its linguistic feature analysis includes characteristics specific to Japanese, such as character types. 
However, other languages, such as English and Chinese, also have unique linguistic features that may influence preferences, such as uppercase usage in English. 
It is important to note that our findings do not necessarily generalize to other languages.
Future work could extend the dataset to multiple languages to explore whether certain linguistic features affecting preferences are shared across languages.
To realize this multilingual extension, there are two possible approaches for multilingual adaptation: translating existing datasets like \dataset~or constructing new ones from scratch. Given that ads often include language- and region-specific proper nouns (e.g., product or service names), translation may lead to unnatural results. Therefore, we believe building datasets from scratch is more appropriate. This would involve collecting ad texts in the target language and applying the same process: paraphrase generation, identification, and preference annotation.

\paragraph{Limited Participants in Preference Judgments}
Due to time and financial constraints, the preference judgments were conducted with ten participants. 
Therefore, their preferences may not accurately reflect those of a broader population. 
To obtain more reliable and robust preference judgment results, collecting opinions from a larger number of participants is necessary.
Additionally, this study recruited only Japanese participants. 
Since preferences can be influenced by demographic factors such as nationality, age, and gender, by collecting such additional information, it would be possible to analyze whether these factors influence preferences. An analysis incorporating demographic information would be a valuable future direction.

\bibliography{main}

\begin{thebibliography}{46}
\providecommand{\natexlab}[1]{#1}

\bibitem[{Bartz et~al.(2008)Bartz, Barr, and Aijaz}]{Bartz2008-ke}
Kevin Bartz, Cory Barr, and Adil Aijaz. 2008.
\newblock \href {https://doi.org/10.1145/1386790.1386792} {Natural language generation for sponsored-search advertisements}.
\newblock In \emph{Proceedings of the 9th ACM Conference on Electronic Commerce}, pages 1--9.

\bibitem[{Brown et~al.(2020)Brown, Mann, Ryder, Subbiah, Kaplan, Dhariwal, Neelakantan, Shyam, Sastry, Askell, Agarwal, Herbert-Voss, Krueger, Henighan, Child, Ramesh, Ziegler, Wu, Winter, Hesse, Chen, Sigler, Litwin, Gray, Chess, Clark, Berner, McCandlish, Radford, Sutskever, and Amodei}]{brown_gpt3_2020}
Tom Brown, Benjamin Mann, Nick Ryder, Melanie Subbiah, Jared~D Kaplan, Prafulla Dhariwal, Arvind Neelakantan, Pranav Shyam, Girish Sastry, Amanda Askell, Sandhini Agarwal, Ariel Herbert-Voss, Gretchen Krueger, Tom Henighan, Rewon Child, Aditya Ramesh, Daniel Ziegler, Jeffrey Wu, Clemens Winter, Chris Hesse, Mark Chen, Eric Sigler, Mateusz Litwin, Scott Gray, Benjamin Chess, Jack Clark, Christopher Berner, Sam McCandlish, Alec Radford, Ilya Sutskever, and Dario Amodei. 2020.
\newblock \href {https://proceedings.neurips.cc/paper_files/paper/2020/file/1457c0d6bfcb4967418bfb8ac142f64a-Paper.pdf} {Language models are few-shot learners}.
\newblock In \emph{Advances in Neural Information Processing Systems 33}, volume~33, pages 1877--1901.

\bibitem[{Cegin et~al.(2023)Cegin, Simko, and Brusilovsky}]{cegin-etal-2023-chatgpt}
Jan Cegin, Jakub Simko, and Peter Brusilovsky. 2023.
\newblock \href {https://doi.org/10.18653/v1/2023.emnlp-main.117} {{C}hat{GPT} to replace crowdsourcing of paraphrases for intent classification: Higher diversity and comparable model robustness}.
\newblock In \emph{Proceedings of the 2023 Conference on Empirical Methods in Natural Language Processing}, pages 1889--1905.

\bibitem[{Cohen(1988)}]{cohen1988statisticalpower}
Jacob Cohen. 1988.
\newblock \emph{Statistical Power Analysis for the Behavioral Sciences}, 2nd edition.
\newblock Lawrence Erlbaum Associates, Hillsdale, NJ.

\bibitem[{Dettmers et~al.(2023)Dettmers, Pagnoni, Holtzman, and Zettlemoyer}]{dettmers2023qlora}
Tim Dettmers, Artidoro Pagnoni, Ari Holtzman, and Luke Zettlemoyer. 2023.
\newblock Qlora: efficient finetuning of quantized llms.
\newblock In \emph{Advances in Neural Information Processing Systems 36}.

\bibitem[{Fleiss et~al.(1971)}]{fleiss1971mns}
J.L. Fleiss et~al. 1971.
\newblock {Measuring nominal scale agreement among many raters}.
\newblock \emph{Psychological Bulletin}, 76(5):378--382.

\bibitem[{Fujii et~al.(2024)Fujii, Nakamura, Loem, Iida, Ohi, Hattori, Shota, Mizuki, Yokota, and Okazaki}]{fujii2024swallow}
Kazuki Fujii, Taishi Nakamura, Mengsay Loem, Hiroki Iida, Masanari Ohi, Kakeru Hattori, Hirai Shota, Sakae Mizuki, Rio Yokota, and Naoaki Okazaki. 2024.
\newblock \href {https://openreview.net/forum?id=TQdd1VhWbe} {Continual pre-training for cross-lingual {LLM} adaptation: Enhancing japanese language capabilities}.
\newblock In \emph{First Conference on Language Modeling}.

\bibitem[{Gu et~al.(2025)Gu, Jiang, Shi, Tan, Zhai, Xu, Li, Shen, Ma, Liu, Wang, and Guo}]{gu2025surveyllmasajudge}
Jiawei Gu, Xuhui Jiang, Zhichao Shi, Hexiang Tan, Xuehao Zhai, Chengjin Xu, Wei Li, Yinghan Shen, Shengjie Ma, Honghao Liu, Yuanzhuo Wang, and Jian Guo. 2025.
\newblock \href {https://arxiv.org/abs/2411.15594} {A survey on llm-as-a-judge}.
\newblock \emph{Preprint}, arXiv:2411.15594.

\bibitem[{Hughes et~al.(2019)Hughes, Chang, and Zhang}]{Hughes2019-sh}
J.~Weston Hughes, Keng-hao Chang, and Ruofei Zhang. 2019.
\newblock \href {https://doi.org/10.1145/3292500.3330754} {Generating better search engine text advertisements with deep reinforcement learning}.
\newblock In \emph{Proceedings of the 25th ACM SIGKDD International Conference on Knowledge Discovery and Data Mining}, pages 2269--2277.

\bibitem[{Ishigami(2024)}]{cyberagent-calm3-22b-chat}
Ryosuke Ishigami. 2024.
\newblock \href {https://huggingface.co/cyberagent/calm3-22b-chat} {cyberagent/calm3-22b-chat}.
\newblock Hugging Face.

\bibitem[{Jha et~al.(2023)Jha, Havens, Dohmann, Trott, and Portes}]{jha2023limit}
Aditi Jha, Sam Havens, Jeremy Dohmann, Alex Trott, and Jacob Portes. 2023.
\newblock Limit: Less is more for instruction tuning across evaluation paradigms.
\newblock \emph{arXiv preprint arXiv:2311.13133}.

\bibitem[{Kajiwara et~al.(2021)Kajiwara, Chu, Takemura, Nakashima, and Nagahara}]{kajiwara-etal-2021-wrime}
Tomoyuki Kajiwara, Chenhui Chu, Noriko Takemura, Yuta Nakashima, and Hajime Nagahara. 2021.
\newblock \href {https://doi.org/10.18653/v1/2021.naacl-main.169} {{WRIME}: A new dataset for emotional intensity estimation with subjective and objective annotations}.
\newblock In \emph{Proceedings of the 2021 Conference of the North American Chapter of the Association for Computational Linguistics: Human Language Technologies}, pages 2095--2104.

\bibitem[{Kamigaito et~al.(2021)Kamigaito, Zhang, Takamura, and Okumura}]{Kamigaito2021-iy}
Hidetaka Kamigaito, Peinan Zhang, Hiroya Takamura, and Manabu Okumura. 2021.
\newblock \href {https://aclanthology.org/2021.naacl-industry.32/} {An empirical study of generating texts for search engine advertising}.
\newblock In \emph{Proceedings of the 2021 Conference of the North American Chapter of the Association for Computational Linguistics: Human Language Technologies: Industry Papers}, pages 255--262.

\bibitem[{Klie et~al.(2024)Klie, Eckart~de Castilho, and Gurevych}]{klie-etal-2024-analyzing}
Jan-Christoph Klie, Richard Eckart~de Castilho, and Iryna Gurevych. 2024.
\newblock \href {https://doi.org/10.1162/coli_a_00516} {Analyzing dataset annotation quality management in the wild}.
\newblock \emph{Computational Linguistics}, 50(3):817--866.

\bibitem[{Lan et~al.(2017)Lan, Qiu, He, and Xu}]{lan-etal-2017-continuously}
Wuwei Lan, Siyu Qiu, Hua He, and Wei Xu. 2017.
\newblock \href {https://doi.org/10.18653/v1/D17-1126} {A continuously growing dataset of sentential paraphrases}.
\newblock In \emph{Proceedings of the 2017 Conference on Empirical Methods in Natural Language Processing}, pages 1224--1234, Copenhagen, Denmark. Association for Computational Linguistics.

\bibitem[{Landis and Koch(1977)}]{richard1977iaa}
J.~Richard Landis and Gary~G. Koch. 1977.
\newblock \href {http://www.jstor.org/stable/2529310} {The measurement of observer agreement for categorical data}.
\newblock \emph{Biometrics}, 33(1):159--174.

\bibitem[{Levenshtein(1966)}]{levenshtein1966binary}
Vladimir~I Levenshtein. 1966.
\newblock Binary codes capable of correcting deletions, insertions and reversals.
\newblock \emph{Soviet Physics Doklady}, 10:707.

\bibitem[{Lin(2004)}]{lin-2004-rouge}
Chin-Yew Lin. 2004.
\newblock \href {https://aclanthology.org/W04-1013} {{ROUGE}: A package for automatic evaluation of summaries}.
\newblock In \emph{Proceedings of the {ACL} Workshop: Text Summarization Branches Out}, pages 74--81.

\bibitem[{Liu et~al.(2023)Liu, Iter, Xu, Wang, Xu, and Zhu}]{liu-etal-2023-g-eval}
Yang Liu, Dan Iter, Yichong Xu, Shuohang Wang, Ruochen Xu, and Chenguang Zhu. 2023.
\newblock \href {https://doi.org/10.18653/v1/2023.emnlp-main.153} {{G}-eval: {NLG} evaluation using gpt-4 with better human alignment}.
\newblock In \emph{Proceedings of the 2023 Conference on Empirical Methods in Natural Language Processing}, pages 2511--2522.

\bibitem[{Maekawa et~al.(2010)Maekawa, Yamazaki, Maruyama, Yamaguchi, Ogura, Kashino, Ogiso, Koiso, and Den}]{maekawa-etal-2010-design}
Kikuo Maekawa, Makoto Yamazaki, Takehiko Maruyama, Masaya Yamaguchi, Hideki Ogura, Wakako Kashino, Toshinobu Ogiso, Hanae Koiso, and Yasuharu Den. 2010.
\newblock \href {http://www.lrec-conf.org/proceedings/lrec2010/pdf/99_Paper.pdf} {Design, compilation, and preliminary analyses of {B}alanced {C}orpus of {C}ontemporary {W}ritten {J}apanese}.
\newblock In \emph{Proceedings of the Seventh International Conference on Language Resources and Evaluation}, pages 1483--1486.

\bibitem[{Mishra et~al.(2020)Mishra, Verma, Zhou, Thadani, and Wang}]{mishra2020refinement}
Shaunak Mishra, Manisha Verma, Yichao Zhou, Kapil Thadani, and Wei Wang. 2020.
\newblock \href {https://doi.org/10.1145/3340531.3412720} {Learning to create better ads: Generation and ranking approaches for ad creative refinement}.
\newblock In \emph{Proceedings of the 29th ACM International Conference on Information and Knowledge Management}, pages 2653--2660.

\bibitem[{Mita et~al.(2024)Mita, Murakami, Kato, and Zhang}]{mita-etal-acl2024-camera}
Masato Mita, Soichiro Murakami, Akihiko Kato, and Peinan Zhang. 2024.
\newblock \href {https://doi.org/10.18653/v1/2024.acl-long.54} {Striking gold in advertising: Standardization and exploration of ad text generation}.
\newblock In \emph{Proceedings of the 62th Annual Meeting of the Association for Computational Linguistics}, pages 955--972.

\bibitem[{Murakami et~al.(2023)Murakami, Hoshino, and Zhang}]{murakami2023atgsurvey}
Soichiro Murakami, Sho Hoshino, and Peinan Zhang. 2023.
\newblock \href {https://arxiv.org/abs/2306.12719} {Natural language generation for advertising: A survey}.
\newblock \emph{Preprint}, arXiv:2306.12719.

\bibitem[{Murakami et~al.(2022)Murakami, Zhang, Hoshino, Kamigaito, Takamura, and Okumura}]{murakami-etal-2022-aspect}
Soichiro Murakami, Peinan Zhang, Sho Hoshino, Hidetaka Kamigaito, Hiroya Takamura, and Manabu Okumura. 2022.
\newblock \href {https://doi.org/10.18653/v1/2022.naacl-industry.9} {Aspect-based analysis of advertising appeals for search engine advertising}.
\newblock In \emph{Proceedings of the 2022 Conference of the North American Chapter of the Association for Computational Linguistics: Human Language Technologies: Industry Track}, pages 69--78.

\bibitem[{Murakami et~al.(2025)Murakami, Zhang, Kamigaito, Takamura, and Okumura}]{murakami2025adparaphrasev1}
Soichiro Murakami, Peinan Zhang, Hidetaka Kamigaito, Hiroya Takamura, and Manabu Okumura. 2025.
\newblock \href {https://arxiv.org/abs/2502.04674} {{A}d{P}araphrase: Paraphrase dataset for analyzing linguistic features toward generating attractive ad texts}.
\newblock \emph{Preprint}, arXiv:2502.04674.

\bibitem[{OpenAI(2024)}]{openai2024gpt4o}
OpenAI. 2024.
\newblock \href {https://openai.com/index/hello-gpt-4o/} {Hello gpt-4o}.
\newblock Accessed: 2025-01-03.

\bibitem[{Papineni et~al.(2002)Papineni, Roukos, Ward, and Zhu}]{papineni-etal-2002-bleu}
Kishore Papineni, Salim Roukos, Todd Ward, and Wei-Jing Zhu. 2002.
\newblock \href {https://aclanthology.org/P02-1040} {{BLEU}: A method for automatic evaluation of machine translation}.
\newblock In \emph{Proceedings of the 40th Annual Meeting of the Association for Computational Linguistics}, pages 311--318.

\bibitem[{Park et~al.(2024)Park, Rafailov, Ermon, and Finn}]{park-etal-2024-disentangling}
Ryan Park, Rafael Rafailov, Stefano Ermon, and Chelsea Finn. 2024.
\newblock \href {https://doi.org/10.18653/v1/2024.findings-acl.297} {Disentangling length from quality in direct preference optimization}.
\newblock In \emph{Findings of the Association for Computational Linguistics: ACL 2024}, pages 4998--5017.

\bibitem[{Pryzant et~al.(2018)Pryzant, Basu, and Sone}]{pryzant-etal-2018-interpretable}
Reid Pryzant, Sugato Basu, and Kazoo Sone. 2018.
\newblock \href {https://doi.org/10.18653/v1/W18-5415} {Interpretable neural architectures for attributing an ad{'}s performance to its writing style}.
\newblock In \emph{Proceedings of the 2018 {EMNLP} Workshop {B}lackbox{NLP}: Analyzing and Interpreting Neural Networks for {NLP}}, pages 125--135.

\bibitem[{Rafailov et~al.(2023)Rafailov, Sharma, Mitchell, Ermon, Manning, and Finn}]{rafailov2024dpo}
Rafael Rafailov, Archit Sharma, Eric Mitchell, Stefano Ermon, Christopher~D. Manning, and Chelsea Finn. 2023.
\newblock Direct preference optimization: your language model is secretly a reward model.
\newblock In \emph{Advances in Neural Information Processing Systems 36}.

\bibitem[{Sato et~al.(2008)Sato, Matsuyoshi, and Kondoh}]{sato-etal-2008-automatic}
Satoshi Sato, Suguru Matsuyoshi, and Yohsuke Kondoh. 2008.
\newblock \href {http://www.lrec-conf.org/proceedings/lrec2008/pdf/165_paper.pdf} {Automatic assessment of {J}apanese text readability based on a textbook corpus}.
\newblock In \emph{Proceedings of the Sixth International Conference on Language Resources and Evaluation}.

\bibitem[{Schwab(2013)}]{schwab2013how-to-write-a-good-advertisement}
Victor~O. Schwab. 2013.
\newblock \emph{How to Write a Good Advertisement: A Short Course in Copywriting}, illustrated edition edition.
\newblock Echo Point Books \& Media.

\bibitem[{Shaib et~al.(2024)Shaib, Elazar, Li, and Wallace}]{shaib-etal-2024-detection}
Chantal Shaib, Yanai Elazar, Junyi~Jessy Li, and Byron~C Wallace. 2024.
\newblock \href {https://doi.org/10.18653/v1/2024.emnlp-main.368} {Detection and measurement of syntactic templates in generated text}.
\newblock In \emph{Proceedings of the 2024 Conference on Empirical Methods in Natural Language Processing}, pages 6416--6431.

\bibitem[{Takaoka et~al.(2018)Takaoka, Hisamoto, Kawahara, Sakamoto, Uchida, and Matsumoto}]{takaoka2018sudachi}
Kazuma Takaoka, Sorami Hisamoto, Noriko Kawahara, Miho Sakamoto, Yoshitaka Uchida, and Yuji Matsumoto. 2018.
\newblock Sudachi: a japanese tokenizer for business.
\newblock In \emph{Proceedings of the Eleventh International Conference on Language Resources and Evaluation}.

\bibitem[{Wang and Pomplun(2012)}]{hsuehcheng_wang__2012}
Hsueh-Cheng Wang and Marc Pomplun. 2012.
\newblock \href {https://doi.org/10.1167/12.6.26} {{The attraction of visual attention to texts in real-world scenes}}.
\newblock \emph{Journal of Vision}, 12(6):26--26.

\bibitem[{Wang et~al.(2024)Wang, Li, Chen, Cai, Zhu, Lin, Cao, Kong, Liu, Liu, and Sui}]{wang-etal-2024-large-language-models-fair}
Peiyi Wang, Lei Li, Liang Chen, Zefan Cai, Dawei Zhu, Binghuai Lin, Yunbo Cao, Lingpeng Kong, Qi~Liu, Tianyu Liu, and Zhifang Sui. 2024.
\newblock \href {https://doi.org/10.18653/v1/2024.acl-long.511} {Large language models are not fair evaluators}.
\newblock In \emph{Proceedings of the 62nd Annual Meeting of the Association for Computational Linguistics (Volume 1: Long Papers)}, pages 9440--9450.

\bibitem[{Wang et~al.(2021)Wang, Gu, Cao, Zhao, Yan, Middha, and Xie}]{Wang2021-uq}
Xiting Wang, Xinwei Gu, Jie Cao, Zihua Zhao, Yulan Yan, Bhuvan Middha, and Xing Xie. 2021.
\newblock \href {https://doi.org/10.1145/3447548.3467105} {Reinforcing pretrained models for generating attractive text advertisements}.
\newblock In \emph{Proceedings of the 27th ACM SIGKDD International Conference on Knowledge Discovery and Data Mining}, pages 3697--3707.

\bibitem[{Wei et~al.(2022)Wei, Bosma, Zhao, Guu, Yu, Lester, Du, Dai, and Le}]{wei2022instructiontuning}
Jason Wei, Maarten~Paul Bosma, Vincent Zhao, Kelvin Guu, Adams~Wei Yu, Brian Lester, Nan Du, Andrew~Mingbo Dai, and Quoc~V. Le. 2022.
\newblock \href {https://openreview.net/forum?id=gEZrGCozdqR} {Finetuned language models are zero-shot learners}.
\newblock In \emph{The Tenth International Conference on Learning Representations}.

\bibitem[{Wolf et~al.(2020)Wolf, Debut, Sanh, Chaumond, Delangue, Moi, Cistac, Rault, Louf, Funtowicz, Davison, Shleifer, von Platen, Ma, Jernite, Plu, Xu, Le~Scao, Gugger, Drame, Lhoest, and Rush}]{wolf-etal-2020-transformers}
Thomas Wolf, Lysandre Debut, Victor Sanh, Julien Chaumond, Clement Delangue, Anthony Moi, Pierric Cistac, Tim Rault, Remi Louf, Morgan Funtowicz, Joe Davison, Sam Shleifer, Patrick von Platen, Clara Ma, Yacine Jernite, Julien Plu, Canwen Xu, Teven Le~Scao, Sylvain Gugger, Mariama Drame, Quentin Lhoest, and Alexander Rush. 2020.
\newblock \href {https://doi.org/10.18653/v1/2020.emnlp-demos.6} {Transformers: State-of-the-art natural language processing}.
\newblock In \emph{Proceedings of the 2020 Conference on Empirical Methods in Natural Language Processing: System Demonstrations}, pages 38--45.

\bibitem[{Yamada et~al.(2020)Yamada, Asai, Shindo, Takeda, and Matsumoto}]{yamada-etal-2020-luke}
Ikuya Yamada, Akari Asai, Hiroyuki Shindo, Hideaki Takeda, and Yuji Matsumoto. 2020.
\newblock \href {https://doi.org/10.18653/v1/2020.emnlp-main.523} {{LUKE}: Deep contextualized entity representations with entity-aware self-attention}.
\newblock In \emph{Proceedings of the 2020 Conference on Empirical Methods in Natural Language Processing (EMNLP)}, pages 6442--6454.

\bibitem[{Youngmann et~al.(2020)Youngmann, Yom-Tov, Gilad-Bachrach, and Karmon}]{youngmann2020}
Brit Youngmann, Elad Yom-Tov, Ran Gilad-Bachrach, and Danny Karmon. 2020.
\newblock \href {https://doi.org/10.1145/3366423.3380211} {The automated copywriter: Algorithmic rephrasing of health-related advertisements to improve their performance}.
\newblock In \emph{Proceedings of The Web Conference 2020}, pages 1366--1377.

\bibitem[{Yuan et~al.(2023)Yuan, Xu, Cao, Zhang, Hui, Li, and Jin}]{yuan2023persuadetoclick}
Yuan Yuan, Fengli Xu, Hancheng Cao, Guozhen Zhang, Pan Hui, Yong Li, and Depeng Jin. 2023.
\newblock \href {https://doi.org/10.1109/TKDE.2021.3110724} {Persuade to click: Context-aware persuasion model for online textual advertisement}.
\newblock \emph{IEEE Transactions on Knowledge and Data Engineering}, 35(2):1938--1951.

\bibitem[{Zhang et~al.(2020)Zhang, Kishore, Wu, Weinberger, and Artzi}]{zhang2020bert-score}
Tianyi Zhang, Varsha Kishore, Felix Wu, Kilian~Q. Weinberger, and Yoav Artzi. 2020.
\newblock \href {https://openreview.net/forum?id=SkeHuCVFDr} {Bertscore: Evaluating text generation with bert}.
\newblock In \emph{The Eighth International Conference on Learning Representations}.

\bibitem[{Zhang et~al.(2019)Zhang, Baldridge, and He}]{zhang-etal-2019-paws}
Yuan Zhang, Jason Baldridge, and Luheng He. 2019.
\newblock \href {https://doi.org/10.18653/v1/N19-1131} {{PAWS}: Paraphrase adversaries from word scrambling}.
\newblock In \emph{Proceedings of the 2019 Conference of the North American Chapter of the Association for Computational Linguistics: Human Language Technologies}, pages 1298--1308.

\bibitem[{Zhou and Bhat(2021)}]{zhou-bhat-2021-paraphrase-survey}
Jianing Zhou and Suma Bhat. 2021.
\newblock \href {https://doi.org/10.18653/v1/2021.emnlp-main.414} {Paraphrase generation: A survey of the state of the art}.
\newblock In \emph{Proceedings of the 2021 Conference on Empirical Methods in Natural Language Processing}, pages 5075--5086. Association for Computational Linguistics.

\bibitem[{Zhu et~al.(2018)Zhu, Lu, Zheng, Guo, Zhang, Wang, and Yu}]{zhu2018selfbleu}
Yaoming Zhu, Sidi Lu, Lei Zheng, Jiaxian Guo, Weinan Zhang, Jun Wang, and Yong Yu. 2018.
\newblock \href {https://doi.org/10.1145/3209978.3210080} {Texygen: A benchmarking platform for text generation models}.
\newblock In \emph{Proceedings of the 41st International ACM SIGIR Conference on Research and Development in Information Retrieval}, pages 1097--1100.

\end{thebibliography}

\clearpage

\appendix

\section{Collecting Paraphrase Candidates\label{appendix:collecting_paraphrase_candidates}}
\dataset~was constructed based on \vone~and CAMERA, a Japanese ad text dataset. Both are governed by the CC BY-NC-SA 4.0 license, and we adhered to the intended use. The details of paraphrase candidate generation using LLMs and crowdworkers are as follows:

\paragraph{LLMs}
\begin{figure}[t]
 \centering
  \includegraphics[width=1\linewidth]{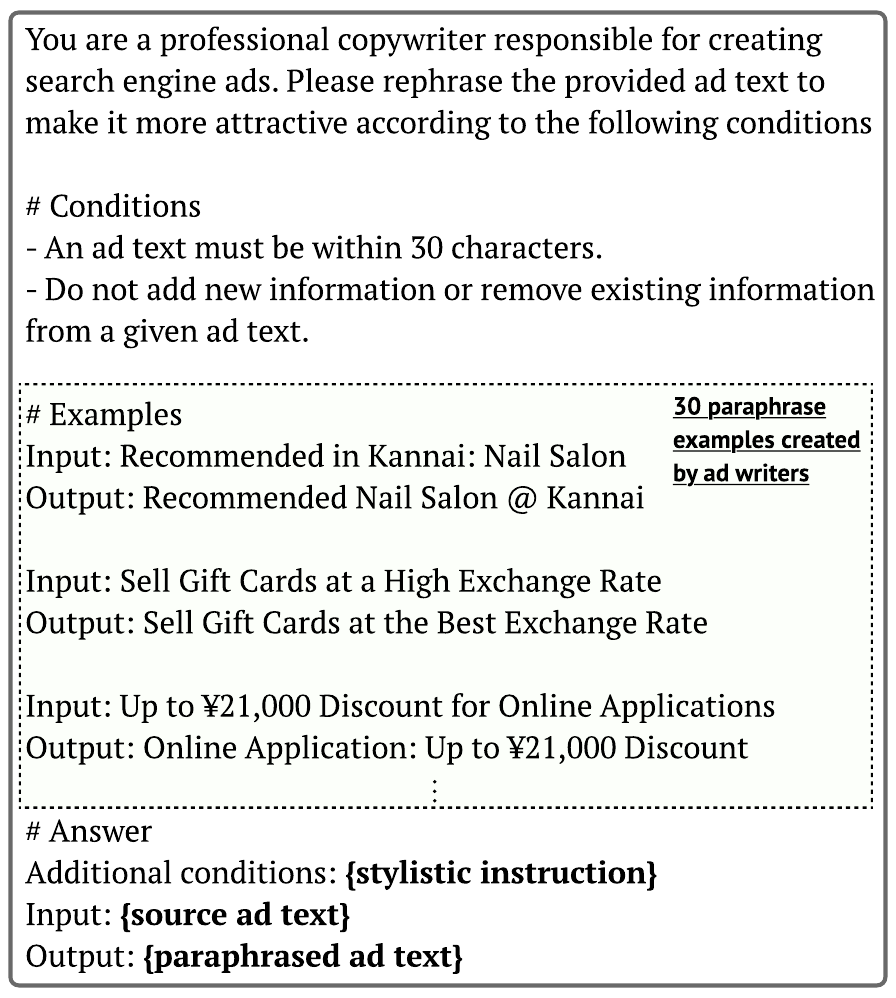}
 \caption{Prompt for paraphrase candidate generation using LLMs.}
 \label{fig:prompt_for_paraphrase_candidate_generation_using_llms}
\end{figure}
\begin{table*}[t]
\centering
{\small
\begin{tabular}{clcl}
\toprule
\multicolumn{1}{c}{\textbf{\#}}  & 
\multicolumn{1}{c}{\textbf{Instructions}} & 
\multicolumn{1}{c}{\textbf{\#}}  & 
\multicolumn{1}{c}{\textbf{Instructions}} \\ \midrule

(1)  & \textit{Use many hiragana characters.           }   & (21) & \textit{Use content words.                            } \\
(2)  & \textit{Use many katakana characters.           }   & (22) & \textit{Use common words.                             } \\
(3)  & \textit{Use many kanji characters.              }   & (23) & \textit{Use technical terms.                          } \\
(4)  & \textit{Write like a news article headline.     }   & (24) & \textit{Use positive words.                           } \\
(5)  & \textit{Use more specific expressions.          }   & (25) & \textit{Use negative words.                           } \\
(6)  & \textit{Use more abstract expressions.          }   & (26) & \textit{Use neutral words.                            } \\
(7)  & \textit{Use first-person pronouns.              }   & (27) & \textit{Use formal language.                          } \\
(8)  & \textit{Use second-person pronouns.             }   & (28) & \textit{Use casual language.                          } \\
(9)  & \textit{Use third-person pronouns.              }   & (29) & \textit{Place important information at the beginning. } \\
(10) & \textit{Use expressions that convey excitement. }   & (30) & \textit{Place important information an the end.       } \\
(11) & \textit{Use expressions that convey joy.        }   & (31) & \textit{Use more complex syntax.                      } \\
(12) & \textit{Use calming and soothing expressions.   }   & (32) & \textit{Use simpler syntax.                           } \\
(13) & \textit{Use expressions that convey urgency.    }   & (33) & \textit{Make it a question.                           } \\
(14) & \textit{Use expressions that encourage action.  }   & (34) & \textit{Use simple words.                             } \\
(15) & \textit{Use brackets.                           }   & (35) & \textit{Use difficult words.                          } \\
(16) & \textit{Use numbers.                            }   & (36) & \textit{Emphasize the benefits.                       } \\
(17) & \textit{Use verbs.                              }   & (37) & \textit{Show how to solve the problem.                } \\
(18) & \textit{Use adjectives.                         }   & (38) & \textit{Include a catchy phrase.                      } \\
(19) & \textit{Use nouns.                              }   & (39) & \textit{Use a visually clear expression.              } \\
(20) & \textit{Use adverbs.                            }   & (40) & \textit{Use an easy-to-read expression.               } \\

\bottomrule
\end{tabular}}
\caption{List of stylistic instructions for paraphrase candidate generation using LLMs}\label{tab:style_list}
\end{table*}

The prompt used to generate the paraphrase candidates is shown in Figure~\ref{fig:prompt_for_paraphrase_candidate_generation_using_llms}. For the few-shot examples, we used 30 paraphrase examples created by professional ad writers from \datasetvone. In addition, to enhance paraphrase diversity, we defined 40 types of stylistic instructions, which are listed in Table \ref{tab:style_list}. These instructions were defined based on previous studies on ATG~\cite{Kamigaito2021-iy} and best practices in copywriting~\cite{schwab2013how-to-write-a-good-advertisement}. During paraphrase generation, a stylistic instruction was randomly selected for each source text. The effectiveness of these stylistic instructions is discussed in Appendix \ref{appendix:effect_of_stylistic_instructions}. For all models, the temperature and top-p were set to 0.8 and 0.95, respectively.

\begin{figure}[t]
 \centering
  \includegraphics[width=1\linewidth]{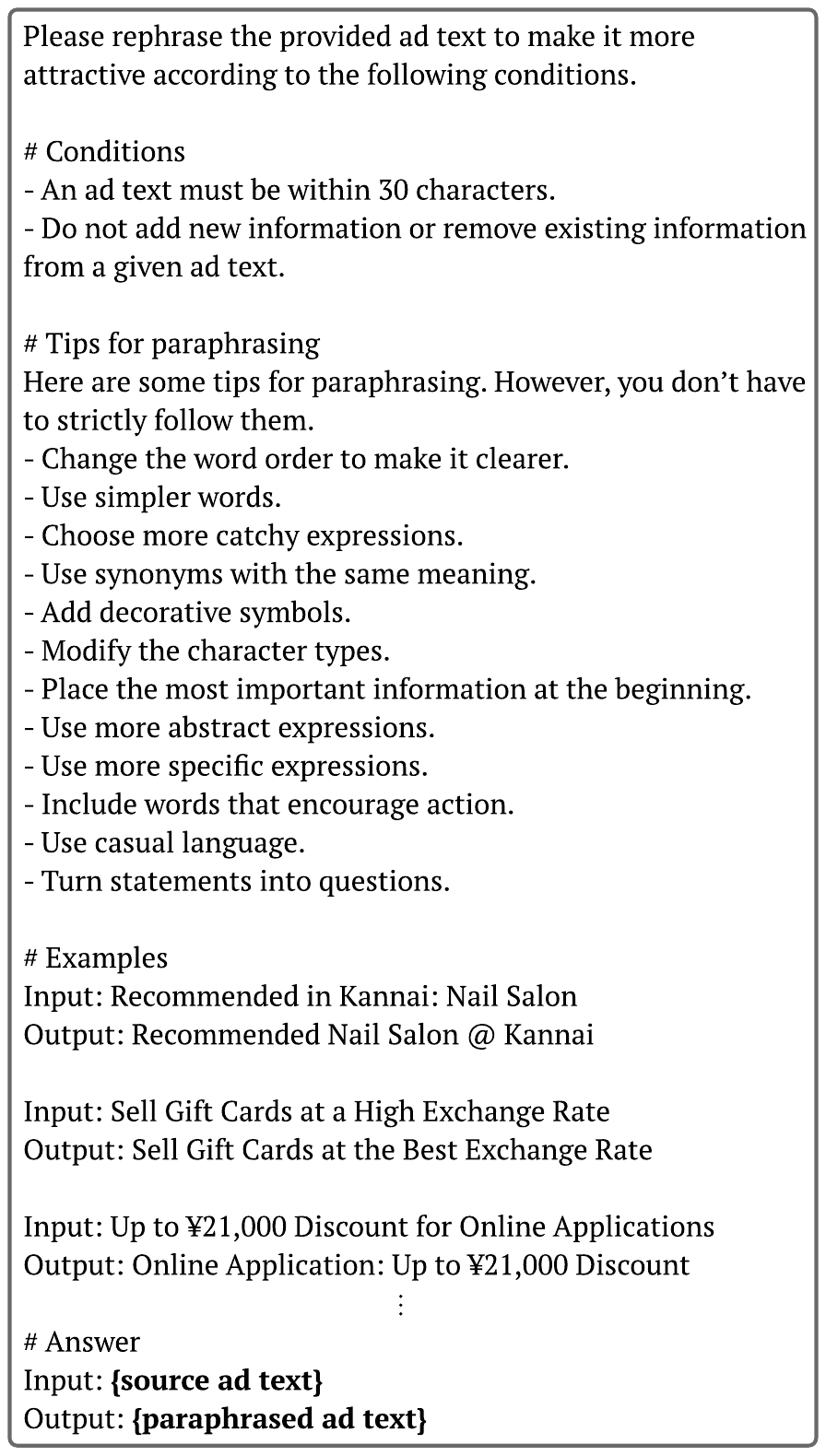}
 \caption{Guidelines for paraphrase candidate creation presented to crowd workers.}
 \label{fig:guidelines_for_paraphrase_candidate_generation_by_crowdworkers}
\end{figure}

\paragraph{Crowdworkers}
Figure \ref{fig:guidelines_for_paraphrase_candidate_generation_by_crowdworkers} shows the annotation guidelines presented to the workers.
The workers were given the same instructions and paraphrasing examples as those provided to the LLMs.
To avoid increasing the annotation burden, we avoided providing explicit stylistic instructions to the human annotators, unlike the method used for LLMs.
Because most workers had no prior experience in ad text creation, the guidelines also included tips on effective paraphrasing.
These guidelines were developed based on insights from previous work \cite{Kamigaito2021-iy} on ATG and best practices in ad text creation \cite{schwab2013how-to-write-a-good-advertisement}.
We used Yahoo! Crowdsourcing as the crowdsourcing platform.\footnote{\url{https://crowdsourcing.yahoo.co.jp/}}
Native Japanese speakers were involved in the annotation process.
Additionally, in accordance with the regulations of the crowdsourcing platform, each worker was compensated with 10 yen per task. The workers were informed in advance that their annotation results would be used for research purposes. Personally identifiable information was not obtained.

\section{Paraphrase Identification\label{appendix:paraphrase_identification}}
\begin{figure}[t]
 \centering
  \includegraphics[width=1\linewidth]{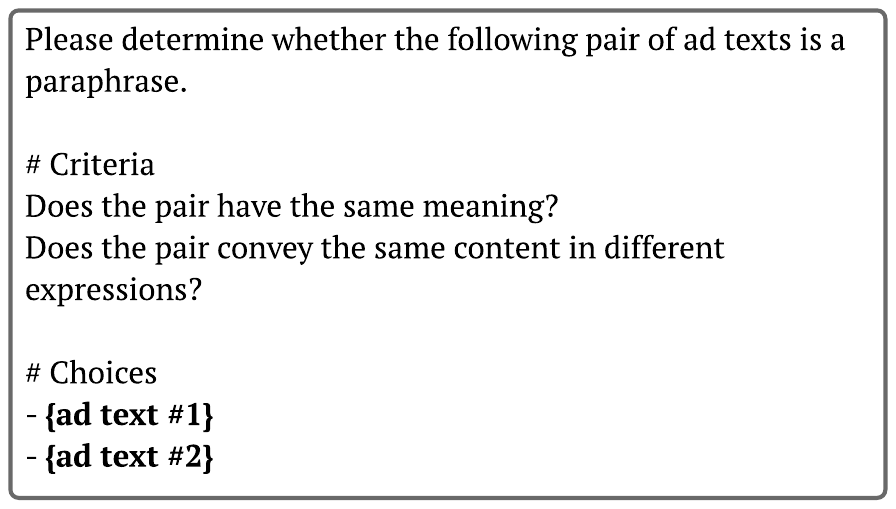}
 \caption{Guidelines for paraphrase identification presented to crowd workers.}
 \label{fig:guidelines_for_paraphrase_identification}
\end{figure}

Figure \ref{fig:guidelines_for_paraphrase_identification} presents the annotation guidelines for paraphrase identification provided to the workers. To ensure consistency between \datasetvone~and \vtwo, we adopted the annotation guidelines used by \citet{murakami2025adparaphrasev1}. The criterion for paraphrase identification is whether two sentences convey the same meaning at the sentence level. We used Yahoo! Crowdsourcing as the crowdsourcing platform. The annotation workers were native Japanese speakers. Each worker was compensated with 10 yen per task in accordance with the regulations of the crowdsourcing platform. No personally identifiable information was collected during the annotation process. The workers were informed that their annotation results would be used for research purposes.

\section{Human Preference Judgments\label{appendix:attractiveness_eval}}
\begin{figure}[t]
 \centering
  \includegraphics[width=1\linewidth]{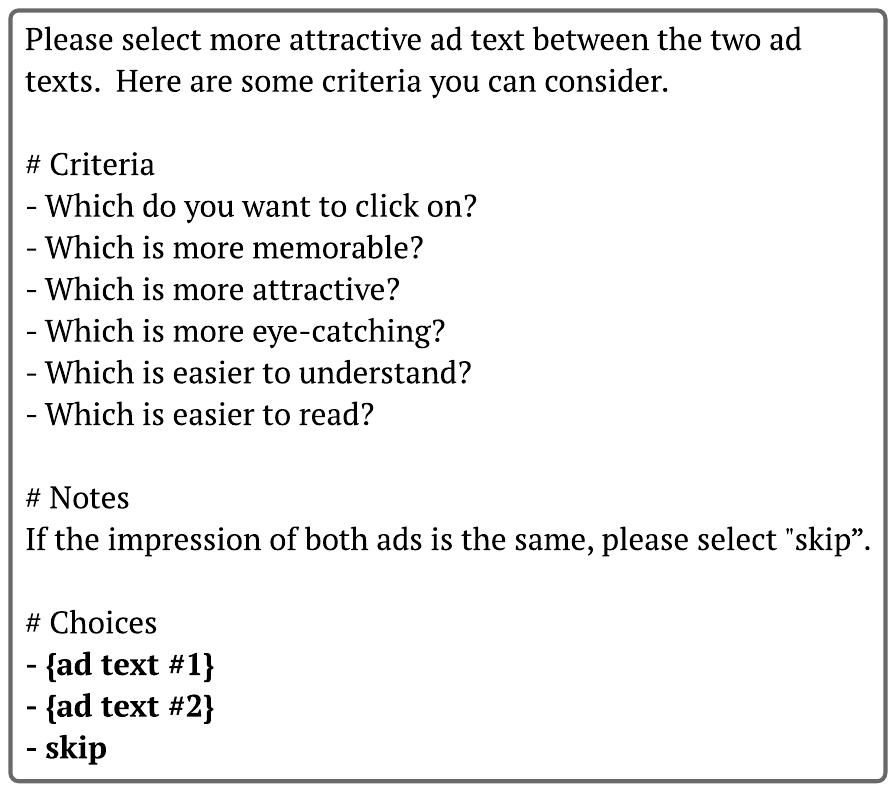}
 \caption{Guidelines for preference judgments presented to crowd workers.}
 \label{fig:guidelines_for_preference_judgments}
\end{figure}

Figure \ref{fig:guidelines_for_preference_judgments} presents the annotation guidelines for the preference judgments provided to the workers. To ensure that preference judgment criteria were consistent with \datasetvone, we adopted the same annotation guidelines as those used by \citet{murakami2025adparaphrasev1}. We used Yahoo! Crowdsourcing as the crowdsourcing platform. Native Japanese speakers were involved in the annotation process. In accordance with the regulations of the crowdsourcing platform, each worker was compensated with 10 yen per task. Personally identifiable information was not obtained. The workers were informed in advance that their annotation results would be used for research purposes.


\section{Example Paraphrase Pairs\label{appendix:example_paraphrase_pairs}}
\begin{table*}[t]
\centering
\setlength{\tabcolsep}{3.5pt}
\label{tab:example_pairs} 
{\small
\begin{tabular}{@{}cll@{}} \toprule
\multicolumn{1}{c}{\textbf{No.}} & 
\multicolumn{1}{c}{\textbf{Ad Text 1}} & 
\multicolumn{1}{c}{\textbf{Ad Text 2}} \\ \midrule

(1) & \textit{Canterbury/Official Online Store} & \textit{Official Online Store of Canterbury} \\
(2) & \textit{Recommended Hair Salon in Tama Center} & \textit{Top Hair Salons -- Tama Center} \\
(3) & \textit{[2022] In-Depth Comparison of No-Installation WiFi} & \textit{2022 Edition: Complete Comparison of No-Installation WiFi} \\
(4) & \textit{Cheap Hotel Reservations in Kumamoto} & \textit{Hotel Reservations in Kumamoto [Affordable]} \\
(5) & \textit{If You're Serious About Solving Hair Problems} & \textit{Hair Problems? Solve Them Seriously} \\
(6) & \textit{[Official] Bleu : Blanc} & \textit{Official Bleu Blanc Website} \\
(7) & \textit{Budget Hotels Near Saiki Station} & \textit{Book Cheap Hotels Near Saiki Station Now} \\
(8) & \textit{[Official] Sumitomo Forestry Detached Homes for Sale} & \textit{[Official] Sumitomo Forestry Residential Homes for Sale} \\
(9) & \textit{Car Appraisal -- Get the Best Price in 32 Seconds} & \textit{What's the Highest Car Appraisal in 32 Seconds?} \\
(10) & \textit{[Fine Save] Official Website} & \textit{[Official] Fine Save Site} \\ \bottomrule

\end{tabular}}
\caption{Example paraphrase pairs in \dataset. \label{tab:list_of_paraphrase_examples}} 
\end{table*}
Table~\ref{tab:list_of_paraphrase_examples} presents the example paraphrase pairs included in \dataset.

\section{Effect of Stylistic Instructions\label{appendix:effect_of_stylistic_instructions}}
\begin{table}[t]
\setlength{\tabcolsep}{2.5pt}
\centering
{\small
\begin{tabular}{@{}l@{\hspace{-2em}}rrr@{}}\toprule
& \multicolumn{1}{c}{\textbf{Levenshtein  ($\downarrow$)}} 
& \multicolumn{2}{c}{\textbf{Self-BLEU ($\downarrow$)}} \\

& \multicolumn{1}{c}{\textbf{edit distance}}
& \multicolumn{1}{c}{\textbf{Word}} &
\multicolumn{1}{c}{\textbf{POS}} \\ \midrule

Baseline prompt                       & 0.542 & 27.8 & 90.3   \\
\hspace{1em} w/ stylistic instruction & \textbf{0.504} & \textbf{26.4} & \textbf{88.9} \\ \bottomrule
\end{tabular}}
\caption{Effects of stylistic instructions on diversity of paraphrase candidates generated by LLMs. \label{tab:diversity_of_paraphrase_candidates}}
\end{table}
We analyzed the effect of stylistic instructions on Principle (2). This analysis evaluated diversity from two perspectives: (1) differences between input and generated texts, and (2) diversity of generated texts. The first perspective uses a similarity score based on the Levenshtein edit distance \cite{levenshtein1966binary}. The second perspective uses self-BLEU \cite{zhu2018selfbleu}. For self-BLEU, we measured lexical and syntactic diversity by evaluating word and part-of-speech (POS) sequences, with POS diversity serving as a proxy for syntactic variation \cite{shaib-etal-2024-detection}.
The results are summarized in Table \ref{tab:diversity_of_paraphrase_candidates}, where lower scores for both perspectives indicate greater diversity. Introducing stylistic instructions improved the diversity of the paraphrase candidates generated by LLMs for both perspectives. These findings suggest that explicitly specifying textual styles in prompts effectively enhances the diversity of the generated texts.

\begin{table}[t]
\setlength{\tabcolsep}{2.8pt}
\centering
{\small
\begin{tabular}{@{}cl@{\hspace{-2.8em}}rrr@{}}
\toprule
  \multicolumn{1}{c}{\multirow{2}{*}{\textbf{\#}}}
& \multicolumn{1}{c}{\textbf{Stylistic}}
& \multicolumn{1}{c}{\textbf{Levenshtein  ($\downarrow$)}} 
& \multicolumn{2}{c}{\textbf{Self-BLEU ($\downarrow$)}} \\

& \multicolumn{1}{c}{\textbf{instruction}}
& \multicolumn{1}{c}{\textbf{edit distance}}
& \multicolumn{1}{c}{\textbf{Word}} 
& \multicolumn{1}{c}{\textbf{POS}} \\ \midrule

(1)  & Hiragana                    & 0.472 & 29.6 & 83.4 \\
(2)  & Katakana                    & 0.553 & 32.0 & 85.7 \\
(3)  & Kanji                       & 0.577 & 31.6 & 87.2 \\
(4)  & News title                  & 0.503 & 34.1 & 92.1 \\
(5)  & Specificity                 & 0.524 & 35.4 & 92.5 \\
(6)  & Abstractness                & 0.368 & 27.4 & 89.1 \\
(7)  & First personal pronoun      & 0.527 & 36.1 & 90.8 \\
(8)  & Second personal pronoun     & 0.450 & 28.9 & 86.8 \\
(9)  & Third personal pronoun      & 0.537 & 34.2 & 88.5 \\
(10) & Excitement                  & 0.445 & 23.0 & 87.7 \\
(11) & Joy                         & 0.430 & 27.8 & 90.4 \\
(12) & Ease                        & 0.495 & 38.1 & 93.5 \\
(13) & Urgency                     & 0.485 & 31.1 & 92.2 \\
(14) & Action                      & 0.423 & 28.3 & 90.9 \\
(15) & Brackets                    & 0.622 & 44.6 & 92.0 \\
(16) & Numbers                     & 0.508 & 33.0 & 90.8 \\
(17) & Verbs                       & 0.523 & 39.6 & 90.9 \\
(18) & Adjectives                  & 0.552 & 35.6 & 88.9 \\
(19) & Nouns                       & 0.580 & 34.3 & 83.2 \\
(20) & Adverbs                     & 0.497 & 32.4 & 89.1 \\
(21) & Content words               & 0.550 & 31.1 & 85.9 \\
(22) & Common words                & 0.544 & 35.6 & 85.5 \\
(23) & Uncommon words              & 0.499 & 28.1 & 85.9 \\
(24) & Positive words              & 0.478 & 32.3 & 89.1 \\
(25) & Negative words              & 0.479 & 26.2 & 86.8 \\
(26) & Neutral words               & 0.520 & 34.9 & 83.0 \\
(27) & Formal words                & 0.534 & 28.8 & 83.2 \\
(28) & Casual words                & 0.438 & 31.6 & 90.1 \\
(29) & Important words to left     & 0.465 & 40.1 & 88.6 \\
(30) & Important words to right    & 0.382 & 41.5 & 88.7 \\
(31) & Complex syntax              & 0.500 & 28.6 & 88.2 \\
(32) & Simple syntax               & 0.549 & 32.5 & \textbf{78.3} \\
(33) & Question                    & 0.524 & 42.1 & 91.4 \\
(34) & Simple words                & 0.498 & 35.4 & 83.7 \\
(35) & Complex words               & 0.456 & 24.3 & 89.0 \\
(36) & Benefits                    & \textbf{0.353} & 23.9 & 92.5 \\
(37) & Solutions                   & 0.435 & 26.3 & 91.4 \\
(38) & Catch-copy                  & 0.392 & \textbf{17.7} & 87.3 \\
(39) & Visibility                  & 0.547 & 35.2 & 88.0 \\
(40) & Readability                 & 0.570 & 39.2 & 86.3 \\ \bottomrule
\end{tabular}}
\caption{Impact of each stylistic instruction on textual diversity. For each metric, the stylistic instruction exhibiting the largest impact is indicated in bold.
\label{tab:effect_of_stylistic_instruction}}
\end{table}

Additionally, we analyzed the impact of each stylistic instruction on textual diversity. The results are summarized in Table \ref{tab:effect_of_stylistic_instruction}, where each instruction number corresponds to an item in the stylistic instructions listed in Table \ref{tab:style_list}.
Our results show that the effects vary according to the instruction type. For example, \textit{``Emphasize the benefits''} and \textit{``Place important information at the end''} improved edit distance, whereas \textit{``Include a catchy phrase''} enhanced lexical diversity and \textit{``Use simpler syntax''} contributed to syntactic diversity.

\section{Analysis of Linguistic Features \label{appendix:linguistic_features}}
\subsection{Linguistic Features}

\begin{table}[!th]
\centering
\setlength{\tabcolsep}{1.9pt}
{\small
\begin{tabular}{@{}c@{\hspace{0.5em}}lrrr@{\hspace{1.1em}}r@{}}
\toprule
\multicolumn{2}{c}{\textbf{Features}} & 
\multicolumn{1}{c}{\textbf{df}} & 
\multicolumn{1}{c}{\textbf{N}} & 
\multicolumn{1}{c}{\textbf{$\chi^2$}} & 
\multicolumn{1}{c}{\textbf{$\phi$}} \\ \midrule 
\multirow{3}{*}{\shortstack{Raw text\\features}}  
    & \textit{Text length} & & & & \\
    & \hspace{1em} character$^{\dag,\ddag,\uparrow,\ast}$      & 1  & 2,925 & 723.8  & 0.497 \\ 
    & \hspace{1em} word$^{\ddag,\uparrow,\ast}$                & 1  & 2,725 & 678.4  & 0.499 \\ 
\midrule
\multirow{16}{*}{\shortstack{Lexical\\features}}  
    & \textit{Content words} & & & & \\
    & \hspace{1em} noun$^{\dag,\ddag,\uparrow,\ast}$           & 1  & 1,406 & 326.6  & 0.482 \\ 
    & \hspace{1em} verb$^{\ddag,\downarrow,\ast}$              & 1  & 535   & 6.9    & 0.114 \\ 
    & \hspace{1em} adjective                                   & 1  & 99    & 0.9    & 0.094 \\ 
    & \hspace{1em} adjectival verb$^{\ddag,\uparrow,\ast}$     & 1  & 105   & 8.0    & 0.276 \\ 
    & \hspace{1em} adverb                                      & 1  & 127   & 0.7    & 0.073 \\ 
    & \textit{Lexical choice} & & & & \\
    & \hspace{1em} word frequency$^{\ddag,\downarrow,\ast}$    & 1  & 2,666 & 70.8   & 0.163 \\ 
    & \hspace{1em} common noun$^{\dag,\ddag,\uparrow,\ast}$    & 1  & 1,397 & 288.1  & 0.454 \\ 
    & \hspace{1em} proper noun$^{\ddag,\uparrow,\ast}$         & 1  & 152   & 7.6    & 0.223 \\ 
    & \textit{Character type} & & & & \\
    & \hspace{1em} hiragana$^{\ddag,\downarrow,\ast}$          & 1  & 2,047 & 23.2   & 0.107 \\ 
    & \hspace{1em} katakana$^{\ddag,\uparrow,\ast}$            & 1  & 601   & 42.6   & 0.266 \\ 
    & \hspace{1em} kanji$^{\ddag,\uparrow,\ast}$               & 1  & 1,503 & 257.7  & 0.414 \\ 
    & \hspace{1em} symbol$^{\ddag,\uparrow,\ast}$              & 1  & 2,332 & 795.9  & 0.584 \\ 
    & \hspace{1em} digits$^{\ddag,\uparrow,\ast}$              & 1  & 66    & 21.0   & 0.564 \\ 
\midrule                                
\multirow{6}{*}{\shortstack{Syntactic\\features}} 
    & \textit{Dependency tree} & & & & \\
    & \hspace{1em} depth$^{\ddag,\downarrow,\ast}$             & 1  & 1,914 & 16.9   & 0.094 \\ 
    & \hspace{1em} length                                      & 1  & 2,349 & 1.9    & 0.028 \\ 
    & \textit{Others} & & & & \\
    & \hspace{1em} noun phrases$^{\dag,\ddag,\uparrow,\ast}$   & 1  & 1,895 & 259.8  & 0.370 \\ 
    & \hspace{1em} perplexity$^{\dag,\ddag,\downarrow,\ast}$   & 1  & 3,570 & 223.3  & 0.250 \\ 
\midrule
\multirow{9}{*}{\shortstack{Stylistic\\features}} 
    & \textit{Emotion} & & & & \\
    & \hspace{1em} joy$^{\ddag,\downarrow,\ast}$               & 1  & 693   & 70.1   & 0.318 \\ 
    & \hspace{1em} anticipation$^{\ddag,\uparrow,\ast}$        & 1  & 683   & 89.3   & 0.362 \\ 
    & \hspace{1em} sadness$^{\ddag,\downarrow,\ast}$           & 1  & 17    & 7.2    & 0.653 \\ 
    & \hspace{1em} surprise                                    & 1  & 28    & 0.2    & 0.083 \\ 
    & \textit{Others} & & & & \\
    & \hspace{1em} specificity$^{\ddag,\uparrow,\ast}$         & 1  & 186   & 116.4  & 0.791 \\ 
    & \hspace{1em} brackets$^{\dag,\ddag,\uparrow,\ast}$       & 1  & 1,667 & 1,372.6& 0.907 \\ 
    & \hspace{1em} question marks                              & 1  & 78    & 1.9    & 0.158 \\ 
\bottomrule                                
\end{tabular}}
\caption{Results of the chi-square test for linguistic features. Df, N, and $\phi$ refer to the degree of freedom and the number of cases, and the measure of effect size, respectively. \ddag~indicates linguistic features, identified in \vtwo, that influence preference judgments, while \dag~denotes those identified in \vone. $\uparrow$ and $\downarrow$ indicate that ad texts with higher and lower feature scores, respectively, are preferred. $\ast$ indicates a significant relationship with human preferences ($p<0.01$).}\label{tab:chi_square_test_full}
\end{table}
Table \ref{tab:chi_square_test_full} presents the 26 linguistic feature types used to analyze the factors influencing preference judgments. The definitions and extraction methods are described below. The extraction methods for each feature followed the same procedure outlined in \citet{murakami2025adparaphrasev1}.

\paragraph{Raw Text Features}
Character and word counts were used as raw text features because they affect the informativeness and readability of the text. Sudachi \cite{takaoka2018sudachi} was employed as a tokenizer for Japanese text.

\paragraph{Lexical Features}
The lexical features included the number of content words, character types, and lexical choices. Content words are indicative of the informativeness of ad texts, whereas character types are associated with readability~\cite{sato-etal-2008-automatic}. The number of content words was counted along with each character type. For the lexical choice, assuming that more frequently used words are preferred, the average word frequency was calculated using a balanced corpus of contemporary written Japanese (BCCWJ) \cite{maekawa-etal-2010-design}. Additionally, the number of common and proper nouns was counted.

\paragraph{Syntactic features}
Syntactic features are measures of the complexity and fluency of ad texts. They include dependency tree depth, length of dependency links, number of noun phrases, and perplexity. Dependency parsing and noun phrase extraction were performed using spaCy with GiNZA~\footnote{\url{https://github.com/megagonlabs/ginza}}. Perplexity was calculated using GPT-2~\footnote{\url{https://huggingface.co/rinna/japanese-gpt2-medium}} trained on web-crawled and Wikipedia corpora. The depth of a dependency tree is the longest path from the root to the leaf node, whereas the length of a dependency link is the number of words between the syntactic head and its dependent.

\paragraph{Stylistic Features}
Stylistic features included emotion, textual specificity, and decorative use of symbols in the text. Following \citet{murakami2025adparaphrasev1}, we assigned emotion and textual-specificity labels to each ad text using external classifiers. In addition to previously studied emotions such as \textit{joy} and \textit{anticipation}, we investigated \textit{sadness} and \textit{surprise}. Details of the classifiers can be found in \S\ref{appendix:external_classifier}. 
Regarding the decorative use of symbols, features such as the presence of brackets and question marks were considered. Brackets were included as features because they are widely used in Japanese ad text to emphasize important information and improve readability. Although question marks have not been studied in previous work, they are frequently used in ad texts to attract people's attention; thus, we introduced them in this study. 

\subsection{External Classifiers\label{appendix:external_classifier}}
The following classifiers were used to assign labels for emotion and textual-specificity to each ad text. The use and construction of the classifiers followed the same procedure as that of \citet{murakami2025adparaphrasev1}.

\paragraph{Emotion}
The LUKE model\footnote{\url{https://huggingface.co/Mizuiro-sakura/luke-japanese-large-sentiment-analysis-wrime}} \cite{yamada-etal-2020-luke}, trained on WRIME \cite{kajiwara-etal-2021-wrime}, a Japanese emotion analysis dataset based on social media text, was used to label the emotions in ad texts. This model is an eight-class classifier that assigns the most appropriate emotion from the following eight categories: \textit{joy, sadness, anticipation, surprise, anger, fear, disgust, and trust}. The classifier achieved an accuracy of 68.6\%.

\paragraph{Textual Specificity}
A specificity classifier was created using GPT-4 via the Azure OpenAI API (\texttt{2024-09-01-preview}) with a few-shot setting. This task was formulated as a three-class classification problem, in which the model compared two ad texts to determine which has higher specificity. If both had equivalent specificity, a label of \textit{``equal''} was output. To evaluate model performance, 100 predictions were randomly sampled and manually evaluated, achieving an accuracy of 88.0\%.

\subsection{Results}
\begin{table}[t]
\centering
{\small
\begin{tabular}{@{ }ccrr@{ }}
\toprule
\multicolumn{2}{c}{\textbf{Longer Ad Texts}} & \multicolumn{1}{c}{Ad1} & \multicolumn{1}{c}{Ad2}\\ \midrule
\textbf{Preferred} & Ad1 & 1,308 & 549  \\
\textbf{Ad Texts}  & Ad2 & 200 & 868  \\ \bottomrule
\end{tabular}}
\caption{Example of Cross-tabulation between human preferences and number of characters in ad texts.\label{table:cross_table_for_char_counts}}
\end{table}
Table \ref{tab:chi_square_test_full} presents the results of the chi-square test for all the features. Several features that were not identified in \vone~\cite{murakami2025adparaphrasev1} were found to be strongly related to preference judgments.
Specifically, \ddag~indicates the linguistic features identified in \vtwo that influenced preference judgments, whereas \dag~denotes those identified in \vone.

In addition, we analyzed the relationship between each feature and human preferences by cross-tabulating feature values. Table~\ref{table:cross_table_for_char_counts} shows the cross-tabulation of preference judgments and text length (in characters) for ad text pairs, where Ad 1 and Ad 2 refer to the source and paraphrased ad text, respectively. In Table~\ref{table:cross_table_for_char_counts}, 1,308 cases were observed, where Ad 1, which has a higher character count, was preferred by the majority of evaluators. We performed this analysis for each feature. 
In Table~\ref{tab:chi_square_test_full}, the $\uparrow$ and $\downarrow$ symbols indicate that ad texts with higher and lower feature scores, respectively, are preferred.
For example, we found that ad texts with the following characteristics are preferred: \textit{longer text}, \textit{more nouns}, \textit{lower dependency tree}, \textit{lower perplexity}, and  \textit{inclusion of brackets}, suggesting that these features are key for enhancing the attractiveness of ad texts.

\section{Ad Text Generation\label{appendix:ad_text_generation}}
    
\paragraph{Implementation Details of ATG Models}
\begin{figure}[t]
 \centering
  \includegraphics[width=1\linewidth]{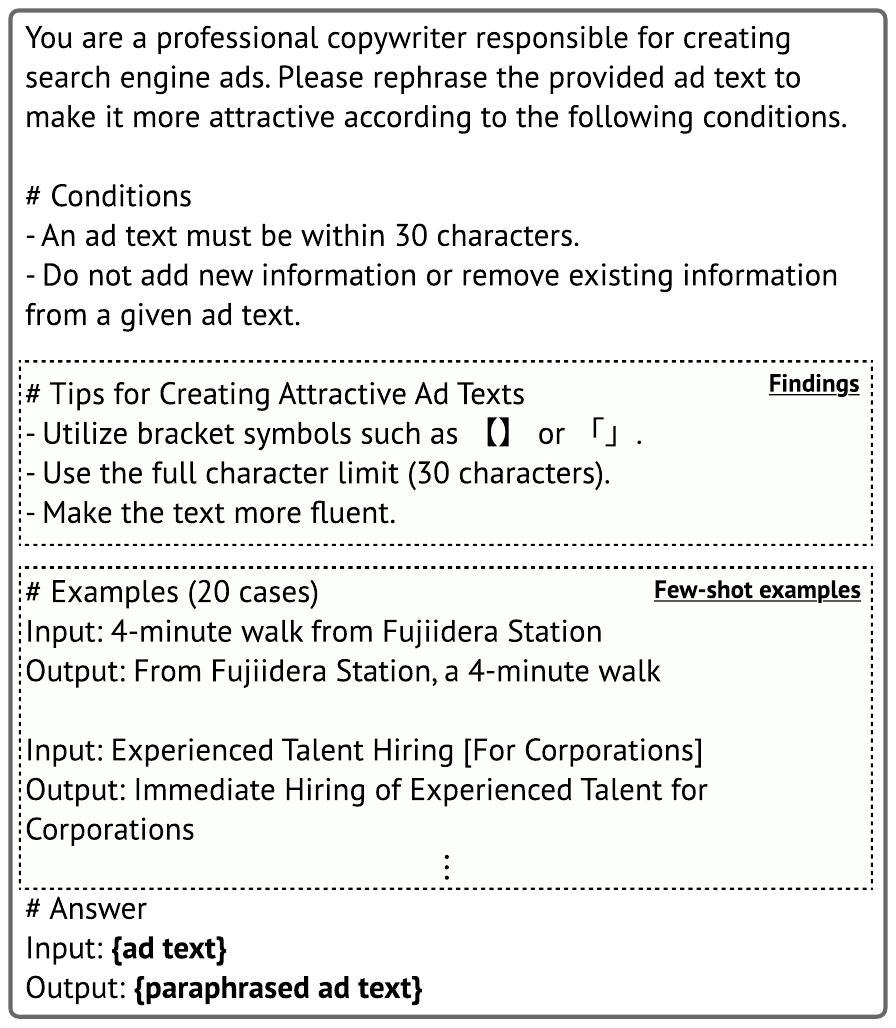}
 \caption{Prompt for \texttt{fewshot-findings} in ATG experiment.}
 \label{fig:prompt_for_atg_experiment}
\end{figure}
Figure \ref{fig:prompt_for_atg_experiment} presents the prompt for the \texttt{fewshot-findings} model, which is an ICL-based approach. The few-shot examples consist of 20 paraphrases randomly sampled from the training data, and the findings incorporate insights from linguistic feature analysis, encouraging longer, more fluent sentences and use of brackets. The \texttt{zeroshot} model relies solely on basic instructions and excludes few-shot examples and findings, whereas the \texttt{zeroshot-findings} model incorporates only the findings.
The instruction-tuned and DPO models were implemented using a quantized low-rank adaptation (QLoRA)~\cite{dettmers2023qlora} and trained for one epoch. The implementation followed the code in the repository\footnote{\url{https://github.com/ghmagazine/llm-book}}.
Greedy decoding was used during inference.

\paragraph{Automatic Evaluation Metrics}
For automatic evaluation, multiple metrics were used, including BLEU (BL) \cite{papineni-etal-2002-bleu}, ROUGE-1 (R-1), ROUGE-2 (R-2), ROUGE-L (R-L) \cite{lin-2004-rouge}, BERTScore (BS) \cite{zhang2020bert-score}, and LLM-based evaluation with GPT-4o \cite{liu-etal-2023-g-eval, gu2025surveyllmasajudge}. These metrics were chosen for their widespread use in paraphrase and other text generation tasks. The F1 scores are reported for ROUGE and BERTScore.
For LLM-based evaluation, we used human evaluation guidelines for PI and preference judgments as prompts. These guidelines are displayed in Figures \ref{fig:guidelines_for_paraphrase_identification} and \ref{fig:guidelines_for_preference_judgments}. The LLM-based evaluation is reference-free, whereas the other metrics are reference-based. For reference-based metrics, the human-created paraphrases in \S\ref{sec:attractive_ad_text_generation} were used as the reference text.

\paragraph{Automatic Evaluation Results}
\begin{table}[t]
\centering
\setlength{\tabcolsep}{2.5pt}
{\small
\begin{tabular}{@{}lrrrrrrr@{}}
\toprule


\multicolumn{1}{c}{\multirow{2}{*}{\textbf{Model}}} &
\multicolumn{1}{c}{\multirow{2}{*}{\textbf{BL}}} &
\multicolumn{1}{c}{\multirow{2}{*}{\textbf{R-1}}} &
\multicolumn{1}{c}{\multirow{2}{*}{\textbf{R-2}}} &
\multicolumn{1}{c}{\multirow{2}{*}{\textbf{R-L}}} &
\multicolumn{1}{c}{\multirow{2}{*}{\textbf{BS}}} &
\multicolumn{2}{c}{\textbf{GPT-4o}} \\  \cmidrule(l){7-8} 
& & & & & &
\multicolumn{1}{c}{\textbf{Para}} &
\multicolumn{1}{c}{\textbf{Att}} \\ \midrule

CALM3-22B & & & & & & & \\
\hspace{0.5em} zeroshot           & 27.4          & 29.3          & 9.8           & 29.2          & 86.5          & 75.0          & 33.0 \\
\hspace{0.5em} zeroshot-findings  & 30.2          & 30.8          & 10.5          & 31.0          & 86.8          & 66.5          & 49.0 \\
\hspace{0.5em} fewshot-findings   & 40.0          & 32.0          & 10.5          & 31.3          & 89.5          & 79.0          & 25.5 \\
\hspace{0.5em} instruct-zeroshot  & \textbf{46.8} & \textbf{32.4} & \textbf{11.0} & \textbf{32.3} & \textbf{90.3} & \textbf{89.5} & 11.0 \\
\hspace{0.5em} dpo-zeroshot       & 15.9          & 29.8          & 10.5          & 29.7          & 81.5          & 59.5          & \textbf{80.0} \\ \midrule
Swallow-70B & & & & & & & \\
\hspace{0.5em} zeroshot           & 41.8          & 31.0          & \textbf{11.5} & 31.2          & 89.2          & 90.0          & 11.5          \\
\hspace{0.5em} zeroshot-findings  & 37.8          & 31.5          & 10.3          & 31.0          & 87.8          & 78.0          & 32.5          \\
\hspace{0.5em} fewshot-findings   & 44.5          & \textbf{32.5} & 10.5          & \textbf{31.6} & 89.2          & 83.0          & 18.0          \\
\hspace{0.5em} instruct-zeroshot  & \textbf{50.5} & 29.4          & 10.7          & 29.0          & \textbf{90.9} & \textbf{90.5} & 6.5           \\
\hspace{0.5em} dpo-zeroshot       & 20.1          & 30.0          & 10.5          & 29.7          & 82.8          & 61.5          & \textbf{65.5} \\ \midrule
GPT-4o & & & & & & & \\
\hspace{0.5em} zeroshot               & 37.7          & 27.1          & 9.2           & 26.9          & 88.1          & 90.0          & 3.0          \\
\hspace{0.5em} zeroshot-findings      & 48.0          & 31.2          & 9.5           & 31.0          & 90.7          & \textbf{92.0} & 5.5          \\
\hspace{0.5em} fewshot-findings       & \textbf{49.5} & \textbf{32.3} & \textbf{10.9} & \textbf{31.5} & \textbf{91.0} & 90.5          & \textbf{7.5} \\ \bottomrule


\end{tabular}}
\caption{Automatic evaluation results of ATG experiment.\label{tab:automatic_evaluation_results_for_atg_experiment}}
\end{table}
Table \ref{tab:automatic_evaluation_results_for_atg_experiment} presents the automatic evaluation results. Here, we report the results of a single run. In automatic evaluations using BLEU, ROUGE, BERTScore, and GPT-4o for PI, the instruction-tuned model and \texttt{fewshot-findings} outperformed the other models. Conversely, in the GPT-4o evaluation of attractiveness, the DPO model achieved the best performance.

\paragraph{Linguistic Features of Generated Texts}
\begin{table}[t]
\setlength{\tabcolsep}{3pt}
\centering

{\small
\begin{tabular}{@{}l@{\hspace{-1em}}rrrr}
\toprule
\multicolumn{1}{c}{\textbf{Model}} & 
\multicolumn{1}{c}{\textbf{Perplexity$\downarrow$}} & 
\multicolumn{1}{c}{\textbf{\#Char$\uparrow$}} & 
\multicolumn{1}{c}{\textbf{Brackets$\uparrow$}} \\ \midrule
CALM3-22B & & & & \\
\hspace{1em} zeroshot          & 155.6          & 27.5           & 5.0            \\
\hspace{1em} zeroshot-findings & 157.6          & 30.7           & 64.5           \\
\hspace{1em} fewshot-findings  & 146.7          & 27.0           & \textbf{69.0}  \\
\hspace{1em} instruct-zeroshot & 168.5          & 24.1           & 48.5           \\
\hspace{1em} dpo-zeroshot      & \textbf{92.2}  & \textbf{42.3}  & 37.0           \\ \midrule
Swallow70B & & & & \\
\hspace{1em}zeroshot           & 158.3          & 27.7           & 13.0           \\
\hspace{1em}zeroshot-findings  & 129.3          & 32.9           & \textbf{89.0}  \\
\hspace{1em}fewshot-findings   & 116.8          & 29.7           & 63.5           \\
\hspace{1em}instruct-zeroshot  & 170.7          & 23.7           & 39.0           \\
\hspace{1em}dpo-zeroshot       & \textbf{70.5}  & \textbf{42.4}  & 42.0           \\ \midrule
GPT-4o & & & & \\
\hspace{1em}zeroshot           & 236.9           & 21.5          & 34.5           \\
\hspace{1em}zeroshot-findings  & 228.9           & 25.0          & \textbf{100.0} \\
\hspace{1em}fewshot-findings   & \textbf{183.8}  & \textbf{25.7} & 73.0           \\ \midrule

Crowdworker                    & 264.3 & 23.8 & 45.8 \\ \midrule
\multicolumn{1}{r}{\textbf{Source ad texts}} & 169.7 & 23.6 & 39.5 \\ \bottomrule
\end{tabular}}
\caption{Linguistic features of generated ad texts. \label{tab:analysis_of_generated_texts_in_atg_experiment}}
\end{table}

 
 
Table~\ref{tab:analysis_of_generated_texts_in_atg_experiment} presents the linguistic features of the generated texts for all models, including PPL, character count, and presence of brackets, which were the key features incorporated into the prompt. These results suggest that the models with higher attractiveness scores in Table~\ref{tab:ad_text_generation_results} performed better across these linguistic features. Notably, DPO-based models exhibited lower PPL and greater character counts, indicating that these factors contribute to the attractiveness of the generated ad texts.

\section{Online Evaluation\label{appendix:online_evaluation}}
In the online evaluation, ad texts paraphrased using the \texttt{few-shot-findings} method described in \S\ref{sec:attractive_ad_text_generation} were deployed to analyze whether paraphrasing more attractive expressions influenced user behavior, such as clicks.
Specifically, an A/B test was conducted to compare an existing ad group as baseline with a paraphrased ad group. This evaluation was conducted using Google Ads.
In search advertising, each ad consisted of 15 headlines and 3 descriptions.
The paraphrasing method was applied to the headlines of the existing ads, whereas the descriptions remained the same as those in the baseline. 
Ads from two companies in the fitness and education industries were used for evaluation, and prior consent was obtained.
The ads from the first company were deployed for two weeks.
For the second company, ads were deployed twice for different durations. The results from the two-week and one-month deployments are reported.

Table~\ref{table:online_evaluation} presents the results of the online evaluation, comparing the click-through rate (CTR), conversion rate (CVR), CTVR, cost per click (CPC), and cost per action (CPA) between the existing and paraphrased ads. Here, CTVR is the product of CTR and CVR and serves as a comprehensive metric for evaluating ad effectiveness.
CPC and CPA represent the costs incurred per click and action, respectively; lower values are preferable for cost efficiency.

Statistical significance tests were conducted using the z-test for CTR, CVR, and CTVR.
For each metric, the z-test compares the rate between the tested ad and baseline, thereby calculating a z-value based on the underlying counts to derive a p-value. The p-values below the significance level (0.01) indicated a statistically significant difference for that metric.

\end{document}